\pdfoutput=1
\documentclass[11pt]{article}

\usepackage[preprint]{acl}

\usepackage{times}
\usepackage{makecell}
\usepackage{latexsym}
\usepackage{graphicx}
\usepackage{amsmath}
\usepackage{booktabs}
\usepackage{tcolorbox}
\usepackage{multirow}
\usepackage{tabularx}
\usepackage{float}
\usepackage{enumitem}

\usepackage[T1]{fontenc}
\usepackage[utf8]{inputenc}

\usepackage{microtype}
\usepackage[T1]{fontenc}

\usepackage{inconsolata}
\usepackage{soul}
\usepackage{adjustbox}
\usepackage{xcolor}
\usepackage[normalem]{ulem}
\useunder{\uline}{\ul}{}
\definecolor{myblue}{RGB}{218,232,252}
\usepackage{colortbl}

\title{SoMeLVLM: A Large Vision Language Model for Social Media Processing}

\author{Xinnong Zhang\textsuperscript{\rm 1}\footnotemark[2],
        Haoyu Kuang\textsuperscript{\rm 2}\footnotemark[2], 
        Xinyi Mou\textsuperscript{\rm 2}, 
        Hanjia Lyu\textsuperscript{\rm 3}, 
        Kun Wu\textsuperscript{\rm 1}, \\
        \textbf{Siming Chen\textsuperscript{\rm 2}, 
        Jiebo Luo\textsuperscript{\rm 3},
        Xuanjing Huang\textsuperscript{\rm 4}, 
        Zhongyu Wei\textsuperscript{\rm 2, \rm 5}\footnotemark[3]
        }
        \\ % [.2cm]
        \normalsize\textsuperscript{\rm 1}{Institute of Science and Technology for Brain-Inspired Intelligence, Fudan University, China} \\
        \normalsize\textsuperscript{\rm 2}{School of Data Science, Fudan University, China} \\
        \normalsize\textsuperscript{\rm 3}{Department of Computer Science, University of Rochester, USA} \\
        \normalsize\textsuperscript{\rm 4}{School of Computer Science, Fudan University, China} \\
        \normalsize\textsuperscript{\rm 5}{Research Institute of Intelligent Complex Systems, Fudan University, China} \\
        \normalsize\texttt{\{xnzhang23, hykuang23, kwu21\}@m.fudan.edu.cn}, \\
        \normalsize\texttt{\{xymou20, simingchen, xjhuang, zywei\}@fudan.edu.cn} \\
        \normalsize\texttt{hlyu5@ur.rochester.edu, jluo@cs.rochester.edu}} 

\begin{document}
\maketitle

% \footnote{* These authors contribute equally to this work.\\}

\begin{abstract}
The growth of social media, characterized by its multimodal nature, has led to the emergence of diverse phenomena and challenges, which calls for an effective approach to uniformly solve automated tasks. The powerful Large Vision Language Models make it possible to handle a variety of tasks simultaneously, but even with carefully designed prompting methods, the general domain models often fall short in aligning with the unique speaking style and context of social media tasks. In this paper, we introduce a Large Vision Language Model for Social Media Processing (SoMeLVLM), which is a cognitive framework equipped with five key capabilities including {\it knowledge \& comprehension}, {\it application}, {\it analysis}, {\it evaluation}, and {\it creation}.  SoMeLVLM is designed to understand and generate realistic social media behavior. We have developed a 654k multimodal social media instruction-tuning dataset to support our cognitive framework and fine-tune our model. Our experiments demonstrate that SoMeLVLM achieves state-of-the-art performance in multiple social media tasks. Further analysis shows its significant advantages over baselines in terms of cognitive abilities.
\end{abstract}

\footnotetext[2]{These authors contribute equally to this work.}
\footnotetext[3]{Corresponding author}
\section{Introduction}

Online social media platforms have been generating an abundance of textual and visual content, offering insights into how individuals communicate, interact, and express themselves. With the advent of communication technology, social media is receiving growing attention as more and more users are active in communities of various topics and interests, which is becoming an important research object as well as a valuable data resource for Computational Social Science (CSS) research~\cite{doi:10.1126/science.aaz8170}. Consequently, automated tasks like sentiment analysis~\cite{carer} and misinformation detection~\cite{gabriel-etal-2022-misinfo} have emerged to help researchers understand social media users and optimize online communities.

\begin{figure}[t!]
	\centering
	\includegraphics[width=1\linewidth]{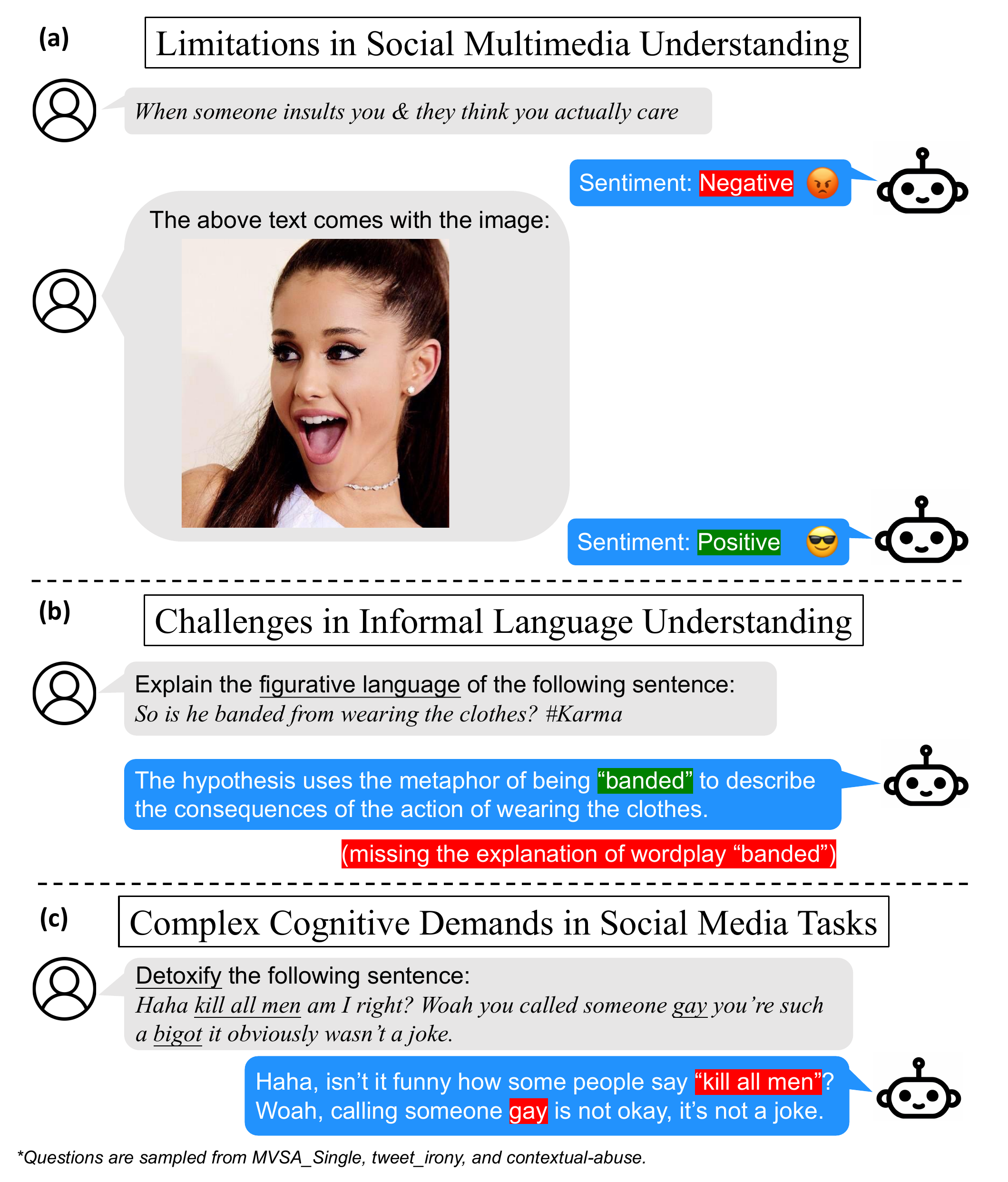}
	\caption{An illustration showing that general domain large language models encounter troubles in (a) social multimedia understanding, (b) informal language understanding, and (c) complex cognitive demands in social media tasks.} 
	\label{fig:intro}
\end{figure}

Recently, Large Language Models (LLMs) and Large Vision Language Models (LVLM)~\citep{openai_2023_chatgpt, chatglm, touvron2023llama2, vicuna2023, lyu2023gpt} have demonstrated their immense capabilities and have offered an effective way to handle automated tasks through prompt engineering. However, research has shown that these generic large models even with extensive prompting practices and evaluations cannot completely replace the traditional research pipeline for CSS, particularly in social media studies~\cite{ziems2023large}. As illustrated in Figure~\ref{fig:intro}, we discover three major challenges faced by general domain models in addressing the nuances of social media: 

\textbf{Limitations in social multimedia understanding.}  General domain LLMs or LVLMs tend to focus more on text over other modalities, which is \textbf{not} consistent with real-world user habits on social media~\cite{liu2023visual, li2023blip2, dai2023instructblip, zhu2023minigpt}. Social media tasks often require fine-grained recognition ability to combine captions and images from a single post and synthesize the user's intention. Genereal domain large models may not possess this level of nuanced multimodal understanding, as shown in Figure~\ref{fig:intro} (a). 

\textbf{Challenges in informal language understanding.} There is a huge gap between the informal speaking style prevalent on social media and the formal language used in other contexts. As a result, general domain LLMs and LVLMs fall short in recognizing sentiment, humor, figurative language, and other related concepts when the sentences are expressed casually. The example shown in Figure~\ref{fig:intro} (b) demonstrates that the model cannot recognize the wordplay ``banded'' in the user's post. 

\textbf{Complex cognitive demands in social media tasks.}
Social media tasks often involve multiple objectives to address high-level social demands that require a combination of complex cognitive abilities and information-processing levels. For instance, the detoxifying task illustrated in Figure~\ref{fig:intro} (c), involves both hate speech detection and content rewriting. However, the models without these abilities struggle to comprehensively address these aspects, resulting in less than satisfactory outputs.

Therefore, to overcome these limitations of the simple prompting strategies and shed light on the investigation of \textit{``how LLMs produce new CSS paradigms built on the multipurpose capabilities of LLMs over the long term''}~\cite{ziems2023large}, we propose \textbf{SoMeLVLM}, a large vision language model tailored for social media processing via extensive and comprehensive supervised fine-tuning. In particular, we establish a solid theoretical foundation. We categorize the tasks concerning social media systematically and build a cognitive pyramid based on Bloom's Taxonomy~\citep{bloom1956taxonomy}, including cognitive levels of \textit{Knowledge \& Comprehension}, \textit{Application}, \textit{Analysis}, \textit{Evaluation}, and \textit{Creation}. These cognitive abilities are derived from different types of users on social media and represent different levels of demands for information processing.

To infuse our model with cognitive abilities, we have curated a large-scale multimodal dataset comprising a total of 654k instances of plain-textual and multimodal data. We then formulate these data into instruction data formats by designing multiple instructional prompts for each task-related subset, covering 12 tasks in total including \textit{emotion}, \textit{humor}, \textit{figurative language}, \textit{hate speech \& toxicity}, \textit{ideology \& stance}, \textit{misinformation}, \textit{trustworthiness \& social bias}, \textit{social factors}, \textit{detoxifying content}, \textit{depolarizing language} \textit{invert opinion}, and \textit{reverse ideology}. Both classification and generative tasks are included in our dataset. 

We apply instruction tuning to our model in two steps. The base language model is tuned initially using textual instruction data, and then a connection module between the vision encoder and the base language model is tuned using multimodal data for advanced cognitive abilities. 

We have conducted both in-domain and out-of-distribution tests on our model and evaluated the performance at both task and cognitive ability levels. The results show that our model effectively overcomes these limitations and achieves state-of-the-art performance in various social media tasks.

To summarize, the main contributions of our paper are as follows:
\begin{itemize}[leftmargin=*]
    \item We propose a large vision language model specifically tailored for social media contexts, capable of delivering high-quality text classification and interpretation under zero-shot conditions, fundamentally simplifying the research workflow in computational social science and improving overall reliability.
    
    \item We construct a comprehensive social media framework by combining cognitive abilities with traditional social media tasks to support different levels of demands in information processing.
    
    \item We contribute to a large-scale, high-quality multimodal social media dataset, encompassing both pure text and multimodal formats, with data from both open-source and self-collected sources, formatted into diverse instruction-tuning formats.
    
\end{itemize}

\begin{figure*}[t!]
	\centering
	\includegraphics[width=1\linewidth]{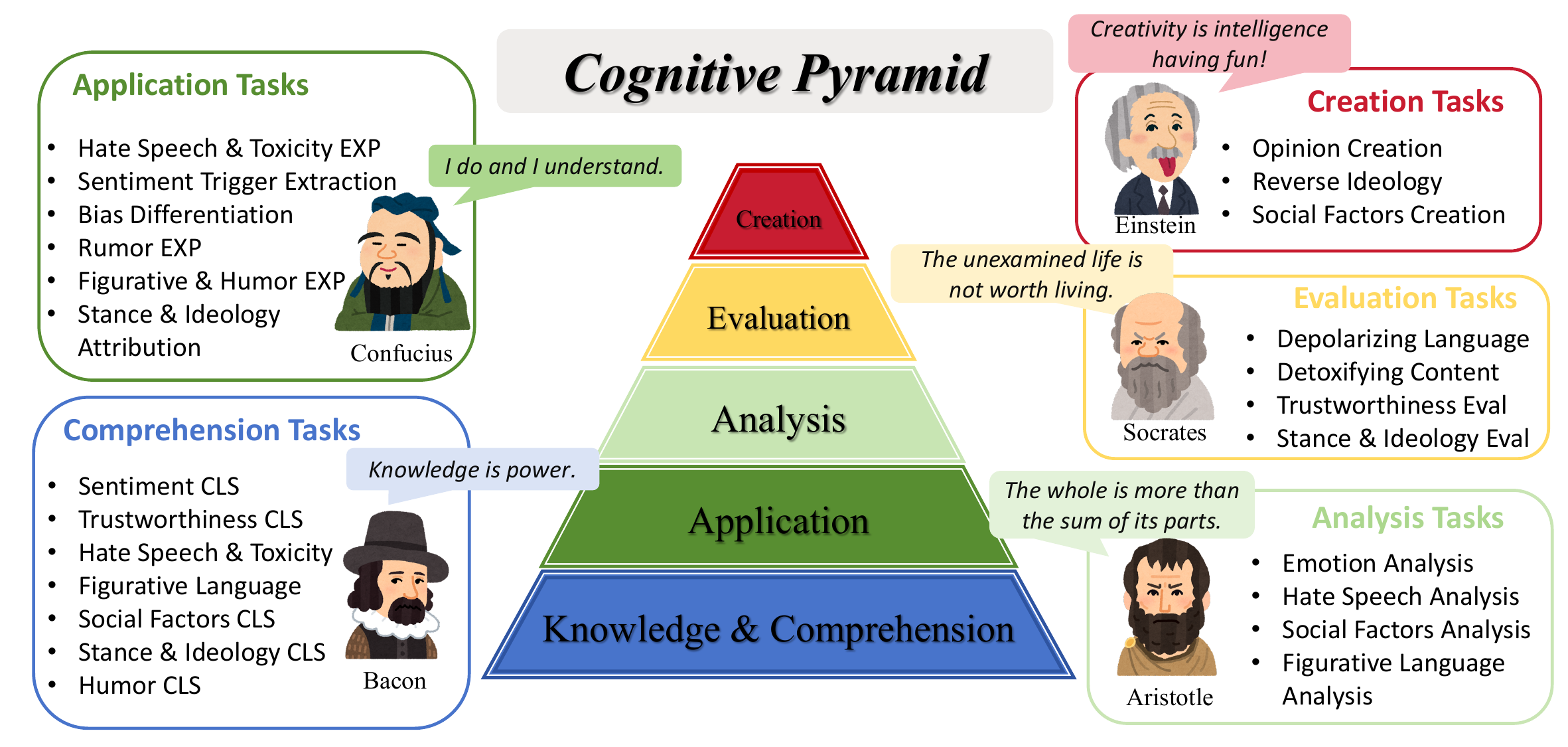}
	\caption{An illustration of the Social Media Cognitive Framework.}
	\label{fig:framework}
\end{figure*}

\section{Related Works}

\subsection{Computational Social Science}\label{review:css}
As an interdisciplinary field, Computational Social Science~\citep{doi:10.1126/science.aaz8170, doi:10.1146/annurev-soc-121919-054621} leverages computational methods to analyze vast datasets, encompassing data from everyday conversations, documents, and books, as well as \textbf{social media content}, to scientifically study linguistic behaviors and social phenomena~\citep{lazer_2, social_phenomena}.

The rise of the Internet has made online interactions a fundamental part of daily life~\citep{doi:10.1146/annurev-soc-071913-043145}, providing invaluable resources for Computational Social Science~\citep{doi:10.1177/0002716215572084}, and paving the way for advancements in social linguistic analysis, such as humor detection~\citep{humor_2}, stance detection~\citep{mou2024}, detection of figurative language~\citep{REYES20121}, and sentiment analysis~\citep{6425642}. Furthermore, it provides guidance for predicting social phenomena, such as fake news detection~\citep{10.1145/3137597.3137600}, the recognition of hate speech~\citep{10.1145/3078714.3078723} and the prediction of ideologies~\citep{mou-etal-2023-uppam}, contributing to a deeper understanding of online and offline social dynamics.

\subsection{Large Vision Language Model}\label{review:lvlm}
The exceptional text understanding and generation capabilities demonstrated by large language models (LLMs)~\citep{openai_2023_chatgpt, touvron2023llama, chatglm, vicuna2023, lyu2023gpt, luo2023valley} have garnered attention across various fields. To further enhance the capability of instruction understanding and generalization ability on unseen datasets, researchers have employed instruction tuning~\cite {wei2022finetuned, chung2022scaling} on LLMs. This approach is capable of augmenting LLMs' comprehension of language within specific domains~\citep{bao2023discmedllm, yue2023disclawllm, chen2023discfinllm}, such as medicine, law, and finance, thereby enhancing performance on related tasks.\par 

By integrating the visual encoders~\citep{pmlr-v139-radford21a, Fang_2023_CVPR} and large language models through linear projection~\citep{NEURIPS2021_01b7575c}, Q-former~\citep{li2023blip2} or cross-attention layers~\citep{NEURIPS2022_960a172b}, LVLMs is capable of addressing a wide range of multimodal tasks. Researchers have also employed instruction tuning on LVLMs, including multitask learning~\citep{pmlr-v139-cho21a}, additional visual components~\citep{li2023blip2}, and instruction-aware components~\citep{dai2023instructblip}. By adopting such an approach, there has indeed been an enhancement in the models' zero-shot generalization capabilities.

% dataset table
\begin{table*}[t!]
    \centering
    \renewcommand\arraystretch{0.85}
    % \resizebox{\textwidth}{!}{%
\begin{tabular}{ccccc}
\toprule[1.1pt]
Level                                      & Category                        & SFT DataSize & Eval Datasize & Total  \\ \midrule
\multirow{8}{*}{\thead{Knowledge \& \\ Comprehension}} & Emotion                         & 63.8k        & 6.5k          & 70.3k  \\
                                            & Humor                           & 18.0k        & 8.3k          & 26.3k  \\
                                            & Figurative Language             & 12.5k        & 4.6k          & 17.1k  \\
                                            & Misinformation                  & 30.4k        & 2.5k          & 32.9k  \\
                                            & Hate Speech \& Toxicity         & 56.5k        & 7.7k          & 64.2k  \\
                                            & Ideology \& Stance              & 25.3k        & 3.8k          & 29.1k  \\
                                            & Trustworthiness \& Social Bias  & 11.0k        & 3.2k          & 14.2k  \\
                                            & Social Factors                 & 55.2k        & 3.5k          & 58.7k  \\ \midrule
\multirow{6}{*}{Application}                   & Emotion                         & 20.0k        & 5.0k          & 25.0k  \\
                                            & Humor                           & 15.0k        & 6.1k          & 21.1k  \\
                                            & Hate Speech \& Toxicity         & 29.6k        & 16.2k         & 45.8k  \\
                                            & Ideology \& Stance              & 4.3k         & 1.0k          & 5.3k   \\
                                            & Trustworthiness \& Social Bias & 30.0k        & -          & 30.0k  \\
                                            & Social Factors                 & 49.0k        & 1.0k          & 50.0k  \\ \midrule
\multirow{4}{*}{Analysis}                   & Figurative Language             & 30.0k        & 2.2k          & 32.2k  \\
                                            & Emotion                         & 18.8k        & 1.5k          & 20.3k  \\
                                            & Hate Speech \& Toxicity         & 12.3k        & 1.5k          & 13.8k  \\
                                            & Social Factors                 & 14.5k        & 0.5k          & 15.0k  \\ \midrule
\multirow{5}{*}{Evaluation}                 & Ideology \& Stance              & 1.3k         & 0.3k          & 1.6k   \\
                                            & Misinformation                  & 8.0k         & 0.5k          & 8.5k   \\
                                            & Trustworthiness \& Social Bias & -         & 0.9k          & 0.9k   \\
                                            & Detoxifying Content             & 25.0k        & 9.9k          & 34.9k  \\
                                            & Depolarizing Language           & 4.3k         & 1.0k          & 5.3k   \\ \midrule
\multirow{3}{*}{Creation}                   & Invert Opinion                  & 1.0k         & -          & 1.0k   \\
                                            & Reverse Ideology                & 4.3k         & 1.0k          & 5.3k   \\
                                            & Social Factors                 & 24.5k          & 0.5k          & 25.0k  \\ \midrule
Total                                       &                                 & 564.6k       & 89.2k         & \textbf{653.8k} \\ \bottomrule[1.1pt]
\end{tabular}%
    % }
    \caption{Composition of data for different cognitive levels}
    \label{tab:datasets}
\end{table*}

\section{Social Media Cognitive Framework}\label{sec:framwork}
In this section, we will present the design of the cognitive pyramid for SoMeLVLM.

\subsection{Framework Design}
To construct a large vision language model capable of understanding and creating multimodal content on social media, we consider concepts from cognitive teaching methods and build a comprehensive multimodal social media cognitive framework, as depicted in Figure~\ref{fig:framework}.
We begin by designing a cognitive pyramid according to Bloom's Taxonomy~\citep{bloom1956taxonomy}, which is a classic teaching theory proposed by Benjamin Bloom in 1956. The pyramid contains five cognitive levels: \textit{Knowledge \& Comprehension}, \textit{Application}, \textit{Analysis}, \textit{Evaluation}, and \textit{Creation}.

We then construct the instruction-tuning data for these five cognitive levels, which is a combination of existing datasets and data collected from social media, resulting in a total of \textbf{654k} instruction pairs. The relation between cognitive levels and different tasks and data statistics are presented in Table~\ref{tab:datasets}.  Each data instance is structured into {\tt text\_input}, {\tt text\_output}, and {\tt image} if it is multimodal, aligning with the format used in Blip2~\cite{li2023blip2}. To ensure the quality of the instruction pairs, we manually design five prompts for \textit{each} dataset. Detailed examples of both plain text and multimodal types are provided in Appendix~\ref{appendix:instruction}.

\subsection{Knowledge \& Comprehension Level}\label{subsec:knowledge}
The Knowledge \& Comprehension level means to recall and understand basic facts. It represents a basic cognitive ability in our framework, which is also the foundation of other higher-level cognitive abilities. Tremendous amounts of concepts are learned via real-world social media data at this level to help the model recognize the content on social media. 

Specifically, the instruction construction of this level consists of various classification tasks within the context of social media, featuring a basic understanding without deeper analysis. 
We have collected a comprehensive collection of open-source datasets annotated by experts in areas such as \textit{Emotion}, \textit{Humor}, \textit{Figurative Language}, \textit{Misinformation}, \textit{Hate speech \& Toxicity}, \textit{Ideology \& Stance}, \textit{Trustworthiness \& Social Bias}, and \textit{Social Factors}. These datasets are structured into question-answering formats, prompting the language model to recognize and categorize these concepts from samples in both textual and multimodal datasets. For binary classification or pairwise choices, a true-or-false question format is applied. For multi-classification, the choices include the entire label space containing up to six candidate answers.

\subsection{Application Level}\label{subsec:applying}
The Application level means to use the information in new situations, which is related to active involvement in social media.
Concepts learned at the former level are used at the application level to explain the phenomena on social media. Consequently, the instruction construction is to make accurate interpretations based on the given ground truth over various social media domains, implying an understanding of the reasons behind the labels.

Given the original ground truth within the datasets annotated by experts, the \texttt{text\_output} of the instruction pair is formulated by appending a concise explanation after the ground truth. Data following the above steps are formulated into tasks including \textit{Emotion Trigger Extraction}, and Interpretation of \textit{Humor}, \textit{Hate Speech}, \textit{Ideology \& Stance}, \textit{Trustworthiness}, and \textit{Social Factors}. For unlabeled data we collect from social media, the ground truth labels are designed as hashtags, personalities, and fields that are closely related to social media. The generated labels along with the explanation are generated by the powerful language model like GPT-4 in advance. To put it briefly, the primary characteristic of the application level is: \textbf{given existing labels}, it enables the model to generate corresponding explanations.

\subsection{Analysis Level}\label{subsec:analysis}
The Analysis level means to draw connections among ideas, which is similar to the application level in that it is a second process based on the concepts learned at the Knowledge \& Comprehension level. The analysis level requires the model to analyze the label and furnish the corresponding interpretations independently. This implies a higher order of capability, enabling it to navigate the rapidly evolving social media landscape. 

We aim for the model to offer explanations \textbf{in the absence of ground truth labels} at this level.
Given the original text or text-image pairs, we provide only the broad context necessary for the analysis of the model such as \textit{Figurative Language Analysis}, \textit{Emotion Analysis} and \textit{Hate Speech Analysis}, and then let the model autonomously generate labels and corresponding explanations. For instance, we instruct the model to analyze the emotional connotation conveyed by the text (or image-text-pair) and elucidate the reasons thereof, while at the application level, we directly present the ground truth emotion and direct the model to analyze the causative factors inducing the said emotion. Therefore, to construct the instruction pairs, the datasets are formulated into a question-answer format, where the question is reformed into a more complex instruction while the answer is generated by GPT-4.

\subsection{Evaluation Level}\label{subsec:evaluation}
The Evaluation level represents the risk forecasting ability, which stands for assessing the probability or likelihood of potential social events and predicting collective trends. At the evaluation level, we pay special attention to the existing prejudices within the data and the abnormal behavior on social media and prompt the model to rewrite original texts or apply knowledge from other domains.

The construction of the data is divided into two aspects. Firstly, for texts that are labeled as containing Hate Speech, we undertake detoxification, and for texts labeled as Liberal or Conservative, we engage in depolarization. Secondly, for texts or text-image pairs labeled as Misinformation, we instruct the model to explain the underlying reasons. Ultimately, the composition of the data is presented in a question-answer format, where the question corresponds to the specific instruction, and the answer is generated by GPT-4.

\subsection{Creation Level}\label{subsec:creation}
The Creation level means to create reliable content related to social media, which is essential during the interaction with the content on social media. This level is considered to be the most complex level. We tackle this demand by setting \textit{reverse} and \textit{creation} tasks, respectively. In the \textit{reverse} task, we require the model to generate opposing viewpoints based on a specified topic and text. In the \textit{create} task, the task is formulated as the generation of new hashtags on social media. 

In terms of instruction construction, regarding the \textit{reverse} task, we formulate the question to prompt the model to generate opposing views on a specific topic, while selecting real statements that hold contrary opinions as the answer. As for the \textit{create} task, we prompt GPT-4 to generate new hashtags related to specific texts, thereby producing question-answer pairs.

\begin{table*}[t!]
\centering
\renewcommand\arraystretch{1.1}
\resizebox{\textwidth}{!}{
\begin{tabular}{@{}lcccccccccccc@{}}
\toprule[1.1pt]
Models          & \multicolumn{2}{c}{\begin{tabular}[c]{@{}c@{}}Hate \\ Speech\end{tabular}} & \multicolumn{2}{c}{\begin{tabular}[c]{@{}c@{}}Misinfor-\\ mation\end{tabular}} & \multicolumn{2}{c}{\begin{tabular}[c]{@{}c@{}}Social \\ Factors\end{tabular}} & \multicolumn{2}{c}{Emotion}     & \multicolumn{2}{c}{Ideology}    & \multicolumn{2}{c}{\begin{tabular}[c]{@{}c@{}}Social Factors \\ OOD\end{tabular}} \\ 
                & Acc*                                 & Acc                                 & Acc*                                   & Acc                                   & Acc*                                  & Acc                                   & Acc*           & Acc            & Acc*           & Acc            & Acc*                                     & Acc                                    \\ \midrule
InstructBlip\(_V\) & 41.62                                & 33.43                               & 47.55                                  & 13.60                                 & 80.02                                 & 40.93                                 & 54.53          & 48.90          & 54.15          & 42.41          & 87.30                                    & 22.59                                  \\
InstructBlip\(_F\) & 50.40                                & 48.43                               & 80.78                                  & 79.00                                 & 81.33                                 & {\ul 73.57}                           & {\ul 58.90}    & {\ul 57.80}    & {\ul 53.69}    & 45.57          & 98.31                                    & {\ul 83.95}                            \\
Blip2           & 52.14                                & {\ul 52.14}                         & 80.60                                  & {\ul 80.60}                           & {\ul 81.83}                           & \textbf{80.89}                        & 57.73          & 57.73          & 53.48          & {\ul 53.48}    & {\ul 99.15}                              & \textbf{95.69}                         \\
Llava           & {\ul 53.35}                          & 9.79                                & \textbf{84.67}                         & 25.40                                 & 72.49                                 & 6.69                                  & 53.39          & 10.10          & 49.79          & 1.58           & 93.75                                    & 3.08                                   \\
MiniGPT4        & 45.12                                & 23.00                               & 65.30                                  & 54.20                                 & 64.08                                 & 36.18                                 & 53.13          & 29.48          & 42.13          & 8.86           & 69.58                                    & 34.29                                  \\
SoMeLVLM        & \textbf{72.57}                       & \textbf{72.57}                      & {\ul 82.60}                            & \textbf{82.60}                        & \textbf{84.07}                        & 67.33                                 & \textbf{63.50} & \textbf{63.47} & \textbf{73.24} & \textbf{55.06} & \textbf{100.00}                          & 61.11                                  \\ \bottomrule[1.1pt]
\end{tabular}
}
\caption{Main results of multimodal classification tasks. We report Acc (overall accuracy) and Acc* (accuracy in instruction-following outputs). The \textbf{bold} number represents the best results, and the \underline{underlined} number represents the second-best results.}
\label{tab:multimodal cls}
\end{table*}

\section{Experimental Setup}
\subsection{Data Split}
After the data construction following the design in \S\ref{sec:framwork}, we fine-tune our model using around 564k training data, which is labeled as \textit{SFT} in Table~\ref{tab:datasets-appendix}. We then evaluate our SoMeLVLM across various aspects of social media, marked as \textit{Eval}, including 14 multimodal datasets and 12 held-out plain text datasets, totaling around 89k data. The specific datasets corresponding to each task and the provided instructions are detailed in the Appendix~\ref{appendix:datasets}.

\subsection{Baseline Models}
For tasks involving plain text, we select Llama-2-7b-chat-hf\citep{touvron2023llama2}, Vicuna-7b-v1.1~\citep{vicuna2023}, and ChatGLM2-6b~\citep{zeng2022glm} as our baseline models. 

For tasks containing images, we choose Blip2~\cite{li2023blip2}, InstructBlip (both Vicuna-based and FlanT5xl-based)~\cite{dai2023instructblip}, Llava~\cite{liu2023visual}, and Minigpt4~\cite{zhu2023minigpt} as our baseline models.

\subsection{Evaluation Metrics}
For \underline{classification (CLS)} tasks, we report the accuracy (Acc) of test results, which involves string matching after proper processing. Specifically, considering the zero-shot setting and the overall instruction-following ability of LVLMs, we report both the accuracy over the whole test set and the accuracy when only valid answers are counted (\textbf{Acc*}). For \underline{generative (GEN)} tasks, we report on automatic metrics such as \textbf{BLEU} and \textbf{ROUGE}. In addition, we employ GPT-4 as a grading assistant through specific prompts to evaluate the test outcomes (\textbf{GPT-Score}). In particular, we task GPT-4 with scoring the model's response on a scale from 0 to 5, where a higher score signifies greater consistency with the ground truth. These prompts can be found in Appendix~\ref{appendix:instruction}.

\subsection{Implementation Details}\label{subsec:implement}
For base language model tuning, we employ the QLoRA method~\cite {dettmers2023qlora} with FastChat~\cite{zheng2023judging}. To tune the connection module, we conduct our experiment following the method of LAVIS~\cite{li-etal-2023-lavis} and choose the connection module of \texttt{blip-vicuna-instruct} as the initial model. Accordingly, the base language model to be fine-tuned is assigned as Vicuna-7b-v1.1. The training and inference process is carried out on eight NVIDIA GeForce RTX3090 and eight RTX4090 GPUs. A mixed precision strategy is employed during the training stage due to the restriction of memory. The base language model is first trained for two epochs with plain text datasets, then the connection module is trained on multimodal datasets for three epochs. In the evaluation stage, we employ \texttt{gpt-4-preview-1106} to output the final score.

% multimodal gen
\begin{table*}[t!]
\centering
\resizebox{\textwidth}{!}{%
\begin{tabular}{llcccccc}
\toprule[1.1pt]
Models                           & Metrics   & \begin{tabular}[c]{@{}c@{}}Hate \\ Speech\end{tabular} & Misinformation & \begin{tabular}[c]{@{}c@{}}Social \\ Factors\end{tabular} & Emotion        & Ideology       & \begin{tabular}[c]{@{}c@{}}Social Factors \\ OOD\end{tabular} \\ \midrule
\multirow{3}{*}{InstructBlip\(_V\)} & BLEU      & {\ul 0.65}                                             & {\ul 1.09}     & {\ul 6.21}                                                & {\ul 0.85}     & 0.60           & 1.14                                                          \\
                                 & ROUGE     & 3.13                                                   & 0.88           & 9.02                                                      & 7.26           & 4.89           & 14.03                                                         \\
                                 & GPT Score & 1.83                                                   & 2.84           & 1.46                                                      & 1.96           & 1.61           & 2.07                                                          \\ \midrule
\multirow{3}{*}{InstructBlip\(_F\)} & BLEU      & 0.24                                                   & 0.05           & 1.16                                                      & 0.28           & 0.78           & 1.51                                                          \\
                                 & ROUGE     & 2.79                                                   & 0.81           & 14.60                                                     & 13.69          & 8.36           & 16.91                                                         \\
                                 & GPT Score & 2.11                                                   & {\ul 2.85}     & {\ul 2.12}                                                & 3.02           & 1.62           & 2.16                                                          \\ \midrule
\multirow{3}{*}{Blip2}           & BLEU      & 0.62                                                   & 0.02           & 0.76                                                      & 0.16           & 0.25           & 0.65                                                          \\
                                 & ROUGE     & 2.25                                                   & 1.89           & 11.99                                                     & {\ul 14.82}    & 4.35           & 12.87                                                         \\
                                 & GPT Score & 1.86                                                   & 2.72           & 1.89                                                      & {\ul 3.08}     & {\ul 2.34}     & 1.61                                                          \\ \midrule
\multirow{3}{*}{Llava}           & BLEU      & 0.36                                                   & 0.00           & 1.89                                                      & 0.64           & {\ul 1.10}     & {\ul 2.29}                                                    \\
                                 & ROUGE     & 4.52                                                   & 0.01           & 12.80                                                     & 5.74           & 8.73           & 20.10                                                         \\
                                 & GPT Score & 1.23                                                   & 0.81           & 1.80                                                      & 1.25           & 1.21           & {\ul 2.27}                                                    \\ \midrule
\multirow{3}{*}{Minigpt4}        & BLEU      & 0.43                                                   & 0.69           & 1.20                                                      & 0.55           & 0.32           & 1.98                                                          \\
                                 & ROUGE     & {\ul 8.84}                                             & {\ul 12.15}    & {\ul 17.20}                                               & 10.81          & {\ul 12.68}    & {\ul 20.73}                                                   \\
                                 & GPT Score & {\ul 2.28}                                             & 2.18           & 1.59                                                      & 2.37           & 1.28           & 1.84                                                          \\ \midrule
 \multirow{3}{*}{SoMeLVLM}            & BLEU      & \textbf{31.04}                                         & \textbf{24.06} & \textbf{14.49}                                            & \textbf{37.65} & \textbf{24.08} & \textbf{10.18}                                                \\
                                 & ROUGE     & \textbf{46.35}                                         & \textbf{43.22} & \textbf{32.87}                                            & \textbf{53.87} & \textbf{41.04} & \textbf{31.03}                                                \\
                                 & GPT Score & \textbf{3.21}                                          & \textbf{2.94}  & \textbf{2.86}                                             & \textbf{3.53}  & \textbf{3.39}  & \textbf{3.45}                                                 \\
                                 \bottomrule[1.1pt]
\end{tabular}
}
\caption{Main results of multimodal generation tasks. We report BLEU-L, ROUGE-L, and GPT Score (0 to 5). The \textbf{bold} number represents the best results, and the \underline{underlined} number represents the second-best results.}
\label{tab:multimodal gen}
\end{table*}

%plain text res
\begin{table*}[t!]
\centering
\renewcommand\arraystretch{1}
% \resizebox{\textwidth}{!}{%
\begin{tabular}{lcccccccc}
\toprule[1.1pt]
\multirow{2}{*}{Models}      & \multirow{2}{*}{Emotion}        &  \multirow{2}{*}{Humor}          & \multirow{2}{*}{\thead{Figurative \\ language}} & \multirow{2}{*}{Misinfo} & \multirow{2}{*}{\thead{Hate \\ Speech}}  & \multirow{2}{*}{Ideology}   & \multirow{2}{*}{Trustworth} & \multirow{2}{*}{\thead{Social \\ Factors}}\\
   &                          &                        &                                      &                                &                           &                                  &                                  \\ 
\midrule
Vicuna & 35.86          & 41.08          & 47.07               & {\ul 59.23}    & 11.94          & 34.15          & 36.60           & 42.68          \\
Llama2 & 40.54          & \textbf{61.31} & {\ul 53.77}         & 41.11          & 12.84          & {\ul 37.77}    & {\ul 59.21}     & 31.61          \\
ChatGLM2       & {\ul 41.20}    & 36.94          & 52.05               & 47.21          & {\ul 14.67}    & 30.07          & \textbf{68.44}  & {\ul 48.23}   \\
SoMeLVLM           & \textbf{80.66} & {\ul 60.47}    & \textbf{61.70}      & \textbf{70.38} & \textbf{22.20} & \textbf{45.23} & 43.52           & \textbf{55.39} \\
\bottomrule[1.1pt]
\end{tabular}%
% }
\caption{Main result of plain text classification tasks under OOD settings; we report Accuracy for these tasks. The \textbf{bold} number represents the best results, and the \underline{underlined} number represents the second-best results.}
\label{table:plain text results 1}
\end{table*}

\begin{table*}[t!]
\centering
\renewcommand\arraystretch{0.95}
\resizebox{\textwidth}{!}{%
\begin{tabular}{llccccccccc}
\toprule[1.1pt]
                                 Models   & Metrics & Emo        & Humor          & Figura     & Hate      & Ideol       & Trust     & Detoxify       & Depolar     & Rever        \\ \midrule
\multirow{3}{*}{Vicuna} & BLEU   & {\ul 7.97}     & {\ul 10.49}    & 8.03           & {\ul 7.01}     & {\ul 9.36}     & {\ul 9.70}      & {\ul 10.43}    & {\ul 22.31}    & {\ul 33.40}     \\
                                & ROUGE  & {\ul 31.31}    & {\ul 36.21}    & {\ul 31.55}    & {\ul 31.24}    & {\ul 32.78}    & 34.13          & {\ul 27.96}    & {\ul 42.72}    & {\ul 51.76}    \\
                                & GPT  & {\ul 3.23}     & {\ul 3.24}     & {\ul 2.57}     & {\ul 3.63}     & {\ul 3.41}     & {\ul 3.13}     & {\ul 2.50}      & {\ul 3.26}     & {\ul 2.98}     \\ \midrule
\multirow{3}{*}{Llama2} & BLEU   & 4.25           & 6.36           & {\ul 10.39}    & 1.79           & 4.75           & 4.73           & 1.31           & 8.40            & 20.54          \\
                                & ROUGE  & 23.50           & 28.37          & 31.32          & 17.41          & 25.01          & 26.54          & 10.94          & 26.72          & 38.06          \\
                                & GPT  & 2.99           & 2.48           & 2.73           & 1.94           & 2.78           & 2.82           & 1.14           & 2.21           & 2.04           \\ \midrule
\multirow{3}{*}{ChatGLM2}       & BLEU   & 6.60            & 8.98           & 7.20            & 4.50            & 6.59           & 9.25           & 6.84           & 13.33          & 21.91          \\
                                & ROUGE  & 29.47          & 34.49          & 29.07          & 28.05          & 29.94          & {\ul 34.35}    & 23.92          & 35.66          & 42.27          \\
                                & GPT  & 3.05           & 2.37           & 2.06           & 2.93           & 2.86           & 2.73           & 2.00              & 2.80            & 2.80            \\ \midrule
\multirow{3}{*}{SoMeLVLM}           & BLEU   & \textbf{26.96} & \textbf{13.81} & \textbf{23.77} & \textbf{17.24} & \textbf{14.60}  & \textbf{12.37} & \textbf{27.13} & \textbf{23.54} & \textbf{44.09} \\
                                & ROUGE  & \textbf{51.88} & \textbf{42.84} & \textbf{45.42} & \textbf{43.10}  & \textbf{39.49} & \textbf{39.06} & \textbf{47.76} & \textbf{45.47} & \textbf{61.96} \\
                                & GPT  & \textbf{3.63}  & \textbf{3.38}  & \textbf{3.02}  & \textbf{3.64}  & \textbf{3.43}  & \textbf{3.59}  & \textbf{2.89}  & \textbf{3.28}  & \textbf{3.41}\\
\bottomrule[1.1pt]
\end{tabular}%
}
\caption{Main result of plain text generative tasks under OOD settings; we report BLEU-L, ROUGE-L, and GPT Score (0 to 5) for these tasks (Hate, Ideol, Trust, Depolar, and Rever denote Hate Speech, Ideology \& Stance, Trustworthiness, Depolarize Language, and Reverse Ideology, respectively.). The \textbf{bold} number represents the best results, and the \underline{underlined} number represents the second-best results.}
\label{table:plain text results 2}
\end{table*}

\section{Results}
\subsection{In-Domain Evaluation}
% \Fixme{Merge Tables 2 and 4, replace color with symbol, Due:2.13}
Given the limited availability of multimodal datasets for social media, we primarily carry out the evaluation of multimodal parts under an in-domain setting. We test our model on 11 datasets across five domains including hate speech, misinformation, social factors, emotion, and ideology. The overall results for classification tasks and generative tasks are shown in Table~\ref{tab:multimodal cls} and Table~\ref{tab:multimodal gen}, respectively. SoMeLVLM has significantly surpassed the baseline LVLMs in all of the five domains in both classification and generative tasks, demonstrating its robust ability to handle a wide range of computational social science tasks.

% Construct according to implementation details
\subsection{Out-of-Distribution Evaluation}
For plain-text parts, we conduct Out-of-Distribution (OOD) evaluation in eleven distinct areas, encompassing emotion, humor, figurative language, hate speech, misinformation, ideology, trustworthiness, social factors, detoxifying content, depolarizing language, and reverse ideology. As shown in Table~\ref{table:plain text results 1} and Table~\ref{table:plain text results 2}, SoMeLVLM achieves new zero-shot SOTA results on all aspects. The OOD evaluation of multimodal parts in the social factors domain involving three custom datasets is also reported as \textit{Social Factor OOD} in Table~\ref{tab:multimodal cls} and Table~\ref{tab:multimodal gen}, which is consistent with the results in the in-domain evaluation.

\subsection{Results Analysis on Cognitive Abilities}
We reform the above results according to the cognitive abilities mentioned in our framework. Specifically, we collect the in-domain performance of multimodal parts (using overall Acc performance) and the OOD performance of plain-text parts at the dataset level and categorize them into \textit{Knowledge \& Comprehension}, \textit{Application}, \textit{Analysis}, \textit{Evaluation}, and \textit{Creation}, five cognitive levels in total. 

\begin{figure}[t!]
    \centering
	\includegraphics[width=\linewidth]{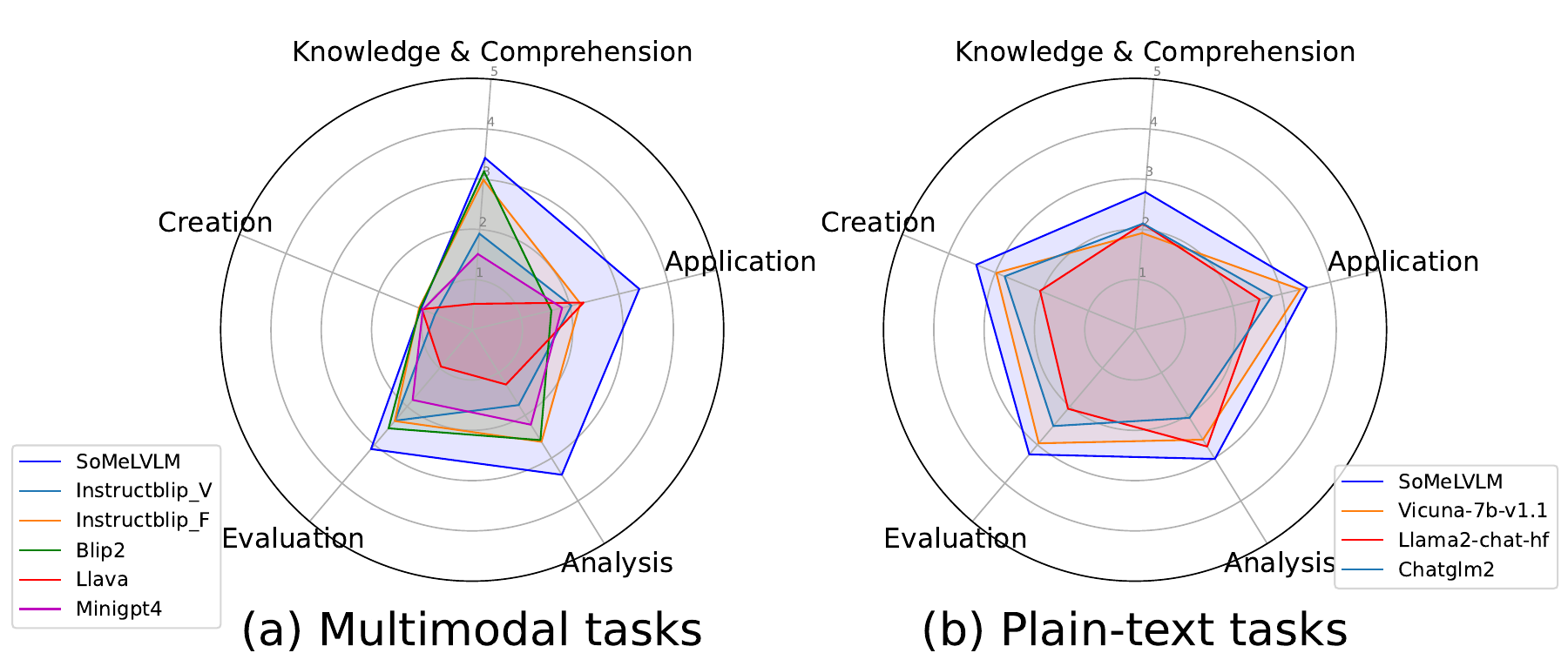}
	\caption{Cognitive abilities performances in (a) Multimodal tasks, and (b) Plain-text tasks.}
	\label{fig:radar}
\end{figure}

The reformed results are shown in Figure~\ref{fig:radar}. Clearly, SoMeLVLM shows greater cognitive ability over baseline models in all of the cognitive levels. At the multimodal \textit{Creation} level, all of the models perform poorly as they are required to generate three hashtags that best describe the post, which is not an easy task even for human beings.

\subsection{Discussion on Instruction Following}
We have noticed that the performance among LVLMs in Table~\ref{tab:multimodal cls} and Table~\ref{tab:multimodal gen} varies significantly, especially for Llava. The overall accuracy of Llava in the classification task is extremely poor, while the accuracy within the valid answer (namely, Acc*) looks good -- even surpassing our model in the misinformation domain. This feeling of separation between Acc and Acc* results from the instruction-following ability of different base language models. When accompanied by the visual information provided by a visual encoder and connection module, base language models of LVLMs at \texttt{7b} level show degeneration in following the output form according to the instructions. Specifically, in our baseline LVLMs, Llama-family (Vicuna-7b-v1.1 and Llama2) base models perform worse than the Flant5-family (Flant5xl) base model. Nevertheless, SoMeLVLM achieves overall the best performance even though we fine-tune it on Vicuna-7b-v1.1, which is the same as InstructBlip$_V$.

Research has found that the ability of instruction-following in LVLMs can be recovered under the few-shot settings~\cite{reform-eval, li2022mvptr}. However in the CSS domain, especially in social media tasks, the zero-shot setting is more proper than a few-shot, as we hope to find a paradigm to handle these tasks automatedly. Besides, in this paper, we want to cultivate complicated cognitive abilities into our model instead of simply emphasizing instruction-following ability, which only belongs to the Knowledge \& Comprehension level.

\section{Conclusion}
In our work, we introduce SoMeLVLM, a multimodal language model for social media processing, wherein we design five cognitive capabilities, each of which is mapped to various levels of social media tasks. 
Building on this, we collect related plain text and multimodal datasets and enhance the capabilities of vision-language models on relevant tasks through instruction tuning. Additionally, we construct an evaluation based on cognitive levels and test our model under zero-shot conditions, comparing it with other advanced LLMs and LVLMs. The experimental results thoroughly demonstrate the superiority of our model. Our work contributes to the computational social science field by providing methods for modeling and evaluating various tasks on social media and a large-scale, high-quality multimodal social media dataset.

\section*{Limitations}
Our work currently focuses on English, and the performances shown in this paper may not be well reproduced in other languages. We are working on a multilingual dataset to improve the robustness under multilingual circumstances.
On the other hand, these neologisms and phrases are often driven by specific cultures, communities, or events, and their meanings may vary across different groups. This suggests that our SoMeLVLM could exhibit interpretive biases towards these terms, especially in the absence of context.

\section*{Ethics Statement}
The data used in this paper are from real users in diverse social media platforms, so the privacy problem is treated cautiously. The data from open-source datasets are safe as the sensitive information has already been masked. For the data we collect, we strictly follow the privacy policy of social media platforms and will carefully avoid personal information before we release our instruction dataset.

\bibliography{preprint}

\appendix

% Dataset table
\begin{table*}[t!]
\centering
% \small
\renewcommand\arraystretch{1.2}
\resizebox{\textwidth}{!}{%
    \begin{tabular}{lllllll|lllllll}
    \hline
    Module                     & Category                       & Dataset                & Size & Task Type & Data Type & Stage    & Module     & Category                       & Dataset                      & Size & Task Type & Data Type & Stage    \\ \hline
    Knowledge \& Comprehension & Emotion                        & Css\_Six\_Emotion      & 30k  & CLS       & Text      & SFT      & Application   & Emotion                        & Css\_Six\_Emotion\_EXP       & 20k  & GEN       & Text      & SFT      \\
    Knowledge \& Comprehension & Emotion                        & Sentiment140           & 15k  & CLS       & Text      & SFT      & Application   & Emotion                        & CARER\_EXP                   & 5K   & GEN       & Text      & Eval     \\
    Knowledge \& Comprehension & Emotion                        & CARER                  & 5k   & CLS       & Text      & Eval     & Application   & Humor                          & humor-pairs\_EXP             & 15k  & GEN       & Text      & SFT      \\
    Knowledge \& Comprehension & Emotion                        & MVSA\_Single           & 2.3k & CLS       & Multi     & SFT/Eval & Application   & Humor                          & hahackathon\#is\_humor\_EXP  & 6.1k & GEN       & Text      & Eval     \\
    Knowledge \& Comprehension & Emotion                        & MVSA\_Multiple         & 8.5k & CLS       & Multi     & SFT/Eval & Application   & Hate Speech \& Toxicity        & jigsaw\_EXP                  & 25k  & GEN       & Text      & SFT      \\
    Knowledge \& Comprehension & Emotion                        & TumEmo                 & 9.5k & CLS       & Multi     & SFT/Eval & Application   & Hate Speech \& Toxicity        & tweet\_offensive\_EXP        & 4.6k & GEN       & Text      & SFT      \\
    Knowledge \& Comprehension & Humor                          & reddit\_jokes          & 4.1k & CLS       & Text      & SFT      & Application   & Hate Speech \& Toxicity        & contextual-abuse\_EXP        & 1.9k & GEN       & Text      & Eval     \\
    Knowledge \& Comprehension & Humor                          & puns                   & 4k   & CLS       & Text      & SFT      & Application   & Hate Speech \& Toxicity        & implicit-hate\_EXP           & 8k   & GEN       & Text      & Eval     \\
    Knowledge \& Comprehension & Humor                          & short\_jokes           & 9.9k & CLS       & Text      & SFT      & Application   & Hate Speech \& Toxicity        & latent\_hatred\_EXP          & 6.3k & GEN       & Text      & Eval     \\
    Knowledge \& Comprehension & Humor                          & hahackathon\#is\_humor & 8.3k & CLS       & Text      & Eval     & Application   & Ideology \& Stance             & ibc\_EXP                     & 4.3k & GEN       & Text      & SFT      \\
    Knowledge \& Comprehension & Figurative Language            & FLUTE                  & 7.5k & CLS       & Text      & SFT      & Application   & Ideology \& Stance             & media\_ideology\_EXP         & 1k   & GEN       & Text      & Eval     \\
    Knowledge \& Comprehension & Figurative Language            & sar                    & 5k   & CLS       & Text      & SFT      & Application   & Trustworthiness \& Social Bias & neutralizing-bias-pairs\_EXP & 30k  & GEN       & Text      & SFT      \\
    Knowledge \& Comprehension & Figurative Language            & tweet\_irony           & 4.6k & CLS       & Text      & Eval     & Application   & Social Factors                & domain\_EXP                  & 25k  & GEN       & Multi     & SFT/Eval \\
    Knowledge \& Comprehension & Misinformation                 & climate\_change        & 24k  & CLS       & Text      & SFT      & Application   & Social Factors                & personality\_EXP             & 25k  & GEN       & Multi     & SFT/Eval \\
    Knowledge \& Comprehension & Misinformation                 & cancer                 & 0.6k & CLS       & Text      & Eval     & Analysis   & Figurative Language            & sar\_EXP                     & 30k  & GEN       & Text      & SFT      \\
    Knowledge \& Comprehension & Misinformation                 & FakeNewsNet            & 6.5k & CLS       & Multi     & SFT/Eval & Analysis   & Figurative Language            & tweet\_irony\_EXP            & 2.2k & GEN       & Text      & Eval     \\
    Knowledge \& Comprehension & Hate Speech \& Toxicity        & jigsaw                 & 30k  & CLS       & Text      & SFT      & Analysis   & Emotion                        & MVSA\_Single\_EXP            & 2.3k & GEN       & Multi     & SFT/Eval \\
    Knowledge \& Comprehension & Hate Speech \& Toxicity        & tweet\_offensive       & 14k  & CLS       & Text      & SFT      & Analysis   & Emotion                        & MVSA\_Multiple\_EXP          & 8.5k & GEN       & Multi     & SFT/Eval \\
    Knowledge \& Comprehension & Hate Speech \& Toxicity        & latent\_hatred         & 6.3k & CLS       & Text      & Eval     & Analysis   & Emotion                        & TumEmo\_EXP                  & 9.5k & GEN       & Multi     & SFT/Eval \\
    Knowledge \& Comprehension & Hate Speech \& Toxicity        & 4chans                 & 2k   & CLS       & Multi     & SFT/Eval & Analysis   & Hate Speech \& Toxicity        & 4chans\_EXP                  & 2k   & GEN       & Multi     & SFT/Eval \\
    Knowledge \& Comprehension & Hate Speech \& Toxicity        & MMHS                   & 7.5k & CLS       & Multi     & SFT/Eval & Analysis   & Hate Speech \& Toxicity        & MMHS\_EXP                    & 7.5k & GEN       & Multi     & SFT/Eval \\
    Knowledge \& Comprehension & Hate Speech \& Toxicity        & hatefulmemes                  & 4.3k & CLS       & Multi     & SFT/Eval & Analysis   & Hate Speech \& Toxicity        & hatefulmemes\_EXP                   & 4.3k & GEN       & Multi     & SFT/Eval \\
    Knowledge \& Comprehension & Ideology \& Stance             & ibc                    & 4.3k & CLS       & Text      & SFT      & Analysis   & Social Factors                & PAN18\_EXP                     & 15k  & GEN       & Multi     & SFT/Eval \\
    Knowledge \& Comprehension & Ideology \& Stance             & vast                   & 18k  & CLS       & Text      & SFT      & Evaluation & Ideology \& Stance             & tweet\_leg\_EXP              & 1k   & GEN       & Multi     & SFT/Eval \\
    Knowledge \& Comprehension & Ideology \& Stance             & election\_stance       & 1.7k & CLS       & Text      & SFT      & Evaluation & Ideology \& Stance             & tweet\_cele\_EXP             & 0.6k & GEN       & Multi     & SFT/Eval \\
    Knowledge \& Comprehension & Ideology \& Stance             & media\_ideology        & 3.5k & CLS       & Text      & Eval     & Evaluation & Misinformation                 & mrf\_headlines\_EXP          & 2k   & GEN       & Text      & SFT      \\
    Knowledge \& Comprehension & Ideology \& Stance             & tweet\_leg             & 1k   & CLS       & Multi     & SFT/Eval & Evaluation & Misinformation                 & FakeNewsNet\_EXP             & 6.5k & GEN       & Multi     & SFT/Eval \\
    Knowledge \& Comprehension & Ideology \& Stance             & tweet\_cele            & 0.6k & CLS       & Multi     & SFT/Eval & Evaluation & Trustworthiness \& Social Bias & rumor\_EXP                   & 0.9k & GEN       & Text      & Eval     \\
    Knowledge \& Comprehension & Trustworthiness \& Social Bias & two-to-lie             & 11k  & CLS       & Text      & SFT      & Evaluation & Detoxifying Content            & jigsaw\_EXP                  & 25k  & GEN       & Text      & SFT      \\
    Knowledge \& Comprehension & Trustworthiness \& Social Bias & hypo-l                 & 3.2k & CLS       & Text      & Eval     & Evaluation & Detoxifying Content            & contextual-abuse\_EXP        & 1.9k & GEN       & Text      & Eval     \\
    Knowledge \& Comprehension & Social Factors                & Stanford Politeness    & 11k  & CLS       & Text      & SFT      & Evaluation & Detoxifying Content            & implicit-hate\_EXP           & 8k   & GEN       & Text      & Eval     \\
    Knowledge \& Comprehension & Social Factors                & complaints             & 3.4k & CLS       & Text      & SFT      & Evaluation & Depolarizing Language          & ibc\_EXP                     & 4.3k & GEN       & Text      & SFT      \\
    Knowledge \& Comprehension & Social Factors                & empathy                & 1.8k & CLS       & Text      & SFT      & Evaluation & Depolarizing Language          & media\_ideology\_EXP         & 1k   & GEN       & Text      & Eval     \\
    Knowledge \& Comprehension & Social Factors                & hayati\_politeness     & 0.3k & CLS       & Text      & Eval     & Creation   & Invert Opinion                 & semeval\_EXP                 & 3k   & GEN       & Text      & SFT      \\
    Knowledge \& Comprehension & Social Factors                & questionintimacy       & 2.2k & CLS       & Text      & Eval     & Creation   & Reverse ideology               & ibc\_EXP                     & 4.3k & GEN       & Text      & SFT      \\
    Knowledge \& Comprehension & Social Factors                & PAN18                    & 15k  & CLS       & Multi     & SFT/Eval & Creation   & Reverse ideology               & media\_ideology\_EXP         & 1k   & GEN       & Text      & Eval     \\
    Knowledge \& Comprehension & Social Factors                & hashtag\_choice        & 25k  & CLS       & Multi     & SFT/Eval & Creation   & Social Factors                & hashtag\_gen\_EXP            & 25k  & GEN       & Multi     & SFT/Eval \\ \hline
    \end{tabular}%
    }
    \caption{Composition of data for different modules}
    \label{tab:datasets-appendix}
\end{table*}

\section{Supplementary on Data Collection and Processing}\label{sec:appendix-data}

\subsection{Datasets}\label{appendix:datasets}
Our datasets come from existing open-source datasets and the raw data we collect. Table~\ref{tab:datasets-appendix} shows all datasets and their relations with cognitive modules and social media tasks. The categories of tasks has been expanded based on the foundation provided by SOCKET\citep{choi2023llms}.

\subsubsection{Existing Datasets}
The following are open-source datasets categorized according to task: \\ 
\textbf{Emotion}\hspace{5px}Binary dataset for coarse-grained sentiment classification: Sentiment140~\citep{go2009twitter}; Multi-class dataset for fine-grained emotion classification: CARER~\citep{carer}. MVSA\_Single and MVSA\_Multiple~\citep{gomez2020mmhs}, TumEmo~\citep{yang2020tumemo}.
\\
\textbf{Humor}\hspace{5px}Binary datasets for humor classification: hahackathon~\citep{meaney-etal-2021-semeval},  reddit\_jokes/puns/short\_jokes~\citep{weller-seppi-2019-humor}, humor-pairs~\citep{hossain-etal-2020-semeval}.
\\
\textbf{Figurative Language}\hspace{5px}Binary datasets for coarse-grained figurative language classification: sar~\citep{khodak-etal-2018-large}; tweet\_irony~\citep{van-hee-etal-2018-semeval}; a multi-class dataset for fine-grained figurative language classification: FLUTE~\citep{chakrabarty-etal-2022-flute}.
\\
\textbf{Misinformation}\hspace{5px}Binary datasets for misinformation classification: climate\_change/cancer~\citep{gabriel-etal-2022-misinfo}, FakeNewsNet~\citep{shu2018fakenewsnet}.
\\
\textbf{Hate Speech \& Toxicity}\hspace{5px}Binary datasets for coarse-grained hate speech classification: implicit-hate~\citep{elsherief-etal-2021-latent}, contextual-abuse~\citep{vidgen-etal-2021-introducing}, tweet\_offensive~\citep{zampieri-etal-2019-semeval}, 4chans~\citep{gonzálezpizarro2022understanding}, memes~\citep{kiela2021hateful}; multi-class datasets for fine-grained hate speech classification: jigsaw~\citep{jigsaw-toxic-comment-classification-challenge}; latent\_hatred~\citep{elsherief-etal-2021-latent}, MMHS~\citep{gomez2020mmhs}.
\\
\textbf{Ideology \& Stance}\hspace{5px}Binary datasets for ideology classification: ibc~\citep{gross2013testing}; Ternary datasets for ideology \& stance classification:  vast~\citep{Allaway2020Zero}; election\_stance~\citep{kawintiranon2021knowledge}; media\_ideology~\citep{baly-etal-2020-detect}, SemEval~\citep{mohammad2016stance}, tweet\_leg~\citep{mou-etal-2021-align}, tweet\_cele~\citep{doi:10.1126/sciadv.abn9418}.
\\
\textbf{Trustworthiness \& Social Bias}\hspace{5px}Binary datasets for trustworthiness classification: two-to-lie~\citep{peskov-etal-2020-takes}; hypo-l~\citep{zhang2022mover}; neutralizing-bias-pairs~\citep{Pryzant_Diehlartinez_Dass_Kurohashi_Jurafsky_Yang_2020}.
\\
\textbf{Social Factors}\hspace{5px}Binary datasets for social factors classification: Stanford Politeness~\citep{fu-etal-2020-facilitating}, complaints~\citep{preotiuc-pietro-etal-2019-automatically}, empathy~\citep{Buechel2018ModelingEA}, hayati\_politeness~\citep{hayati-etal-2021-bert}; Multi-class datasets for social factor classification: questionintimacy~\citep{pei-jurgens-2020-quantifying}, pan~\citep{Pardo2018OverviewOT}.

\subsubsection{Raw Data Collection}
We collect raw social media data with the help of previous related work~\cite{kim2020multimodal}. We then divide these raw data into the following datasets: hashtag\_gen\, hashtag\_choice, domain\_explain, and personality\_explain, each of which contains around 25k data. The ground truths of these datasets are generated by GPT-4V.

\subsection{Instruction Construction}\label{appendix:instruction}
In this section, we will introduce the construction of instructional datasets for various tasks across modules. Specifically, we design a diverse array of prompts manually based on the collected dataset.
% how to construct on each task; prompt intro; distribution intro; example needed
\subsubsection{Knowledge \& Comprehension Module}
As discussed in~\S\ref{subsec:knowledge}, the Knowledge \& Comprehension Module primarily encompasses classification tasks, for which we adapt different prompts to suit the various types of tasks. \\
\textbf{Emotion}\hspace{5px}There are two types of emotion classification tasks: coarse-grained emotion classification, which primarily involves determining whether a statement conveys a positive or negative sentiment, and fine-grained emotion classification, which entails identifying the presence of a specific emotion within a given statement. \par 
\begin{center}
\begin{tcolorbox}[colback=orange!5!white,colframe=orange!55!black,width=0.95\columnwidth,title={Emotion Classification}]
% \small
Determine the emotion conveyed in the text following [Original Text], classifying it as either sadness, joy, love, anger, fear, or surprise. \\
\text{[Original Text]: !<INPUT 0>!} \\
Constraint: Provide a one-word answer. 
\end{tcolorbox}
\label{tab:prompt_emotion_cls}
\end{center}

\begin{center}
\begin{tcolorbox}[colback=red!5!white,colframe=red!55!black,width=0.95\columnwidth,title={Multimodal Emotion Classification}]
This image is associated with the following caption: !<INPUT 0>!.

What sentiment does this combination convey? Positive, neutral, or negative? This is for research purposes.

CONSTRAINTS: only output one word from [positive, neutral, negative].
\end{tcolorbox}
\label{tab:prompt_emotion_cls_multi}
\end{center}
% \begin{table}[H]
% 	\centering
%     \begin{tabularx}{0.9\columnwidth}{ >{\raggedright\arraybackslash}X }
%     \toprule
%     \textbf{Task Description} \\
%     \midrule
%     Determine the emotion conveyed in the text following [Original Text], classifying it as either sadness, joy, love, anger, fear, or surprise. \\
%     \text{[Original Text]: !<INPUT 0>!} \\
%     Constraint: Provide a one-word answer. \\
%     \bottomrule
%     \end{tabularx}
%     \label{tab:prompt_emotion_cls}
% \end{table}
\textbf{Humor}\hspace{5px}The classification of humor is a binary classification task, which involves determining whether a given text is categorized as humor or not humor based on its content.
\begin{center}
\begin{tcolorbox}[colback=orange!5!white,colframe=orange!55!black,width=0.95\columnwidth,title={Humor Classification}]
Assess the provided [Original Text] to determine if it can be categorized as 'humor' or 'not humor'. \\
\text{[Original Text]: !<INPUT 0>!} \\ 
Constraint: Deliver a succinct evaluation, selecting either 'humor' or 'not humor'. 
\end{tcolorbox}
\label{tab:prompt_humor_cls}
\end{center}
% \begin{table}[H]
% 	\centering
%     \begin{tabularx}{0.9\columnwidth}{X}{ >{\raggedright\arraybackslash}X }
%     \toprule
%     \textbf{Task Description} \\
%     \midrule
%     Assess the provided [Original Text] to determine if it can be categorized as 'humor' or 'not humor'. \\
%     \text{[Original Text]: !<INPUT 0>!} \\ 
%     Constraint: Deliver a succinct evaluation, selecting either 'humor' or 'not humor'. \\
%     \bottomrule
%     \end{tabularx}
%     \label{tab:prompt_humor_cls}
% \end{table}
% \par 
\textbf{Figurative Language}\hspace{5px}The classification task of figurative language is twofold: the first type is coarse classification, which determines whether the text contains figurative language, and the second type is fine classification, which identifies the specific type of figurative language used in the text.
\begin{center}
\begin{tcolorbox}[colback=orange!5!white,colframe=orange!55!black,width=0.95\columnwidth,title={Figurative Language Classification}]
Examine the text following [Original Text] for sarcasm. If the meaning contrasts with its literal interpretation, involves a situation of appearance versus reality, or carries a sarcastic tone, classify it as sarcasm. Otherwise, designate it as not-sarcasm.\\
\text{[Original Text]: !<INPUT 0>!}\\
Constraint: Provide a single-word response.
\end{tcolorbox}
\label{tab:prompt_figurative1_cls}
\end{center}

\begin{center}
\begin{tcolorbox}[colback=orange!5!white,colframe=orange!55!black,width=0.95\columnwidth,title={Figurative Language Classification}]
Analyze the [premise] to identify if the [hypothesis] represents sarcasm, creative paraphrase, metaphor, idiom, or simile.\\
\text{[hypothesis]: !<INPUT 0>!} \\
\text{[premise]: !<INPUT 1>!} \\ 
constraint: Single word answer
\end{tcolorbox}
\label{tab:prompt_figurative2_cls}
\end{center}
% \begin{table}[H]
% 	\centering
%     \begin{tabularx}{0.9\columnwidth}{X}{ >{\raggedright\arraybackslash}X }
%     \toprule
%     \textbf{Task Description} \\
%     \midrule
%     Examine the text following [Original Text] for sarcasm. If the meaning contrasts with its literal interpretation, involves a situation of appearance versus reality, or carries a sarcastic tone, classify it as sarcasm. Otherwise, designate it as not-sarcasm.\\
%     \text{[Original Text]: !<INPUT 0>!}\\
%     Constraint: Provide a single-word response.\\
%     \bottomrule
%     \end{tabularx}
%     \label{tab:prompt_figurative_cls}
% \end{table}
\textbf{Misinformation}\hspace{5px}The classification task of misinformation primarily involves identifying given news headlines or text-image pairs, determining whether they represent true information or false information. 
\begin{center}
\begin{tcolorbox}[colback=orange!5!white,colframe=orange!55!black,width=0.95\columnwidth,title={Misinformation Classification}]
Appraise whether the given article [headline] is perceived as misinformation or real based on the \text{[writer\_intent]} and \text{[reader\_action].} \\
\text{[writer\_intent]: !<INPUT 0>!} \\
\text{[reader\_action]: !<INPUT 1>!} \\
\text{[headline]: !<INPUT 2>!} \\
Please answer with single word.
\end{tcolorbox}
\label{tab:prompt_misinfo_cls}
\end{center}

\begin{center}
\begin{tcolorbox}[colback=red!5!white,colframe=red!55!black,width=0.95\columnwidth,title={Multimodal Misinformation Classification}]
This news including an image is associated with the following caption:\\
!<INPUT 0>!\\
Question: Is this news true or false?\\
Choices: [true, false]
\end{tcolorbox}
\label{tab:prompt_misinfo_cls_multi}
\end{center}
% \begin{table}[H]
% 	\centering
%     \begin{tabularx}{0.9\columnwidth}{ >{\raggedright\arraybackslash}X }
%     \toprule
%     \textbf{Task Description} \\
%     \midrule
%     Appraise whether the given article [headline] is perceived as misinformation or real based on the [writer_intent] and [reader_action].\\
%     \text{[writer_intent]: !<INPUT 0>!}\\
%     \text{[reader_action]: !<INPUT 1>!}\\
%     \text{[headline]: !<INPUT 2>!}\\
%     Please answer with single word.\\
%     \bottomrule
%     \end{tabularx}
%     \label{tab:prompt_misinfo_cls}
% \end{table}
\textbf{Hate Speech \& Toxicity}\hspace{5px}The classification task of Hate Speech \& Toxicity is bifurcated into two categories: coarse classification, which determines whether a given text or text-image pair is offensive, and fine classification, which identifies the specific type of hate speech classification. 
\begin{center}
\begin{tcolorbox}[colback=orange!5!white,colframe=orange!55!black,width=0.95\columnwidth,title={Hate Speech Classification}]
\text{[Original text]: !<INPUT 0>!} \\
Evaluate the given [Original Text] to ascertain whether its content should be classified as 'offensive' or 'not offensive'.\\
\text{[Original Text]: !<INPUT 0>!}\\
Restriction: Provide a concise response, choosing either 'offensive' or 'not offensive'.
\end{tcolorbox}
\label{tab:prompt_hate1_cls}
\end{center}

\begin{center}
\begin{tcolorbox}[colback=orange!5!white,colframe=orange!55!black,width=0.95\columnwidth,title={Hate Speech Classification}]
\text{[Original text]: !<INPUT 0>!} \\
Identify the type of hate speech in the text following [original text], labeling it as either white-grievance, threatening, inferiority, stereotypical, incitement irony or other.\\
Restriction: Use only one word for your response.
\end{tcolorbox}
\label{tab:prompt_hate2_cls}
\end{center}

\begin{center}
\begin{tcolorbox}[colback=red!5!white,colframe=red!55!black,width=0.95\columnwidth,title={Multimodal Hate Speech Classification}]
This image is associated with the following caption: '!<INPUT 0>!'.\\
Does this combination exhibit any elements of hate speech?\\
Choices: [true, false]
\end{tcolorbox}
\label{tab:prompt_hate1_cls_multi}
\end{center}

\begin{center}
\begin{tcolorbox}[colback=red!5!white,colframe=red!55!black,width=0.95\columnwidth,title={Multimodal Hate Speech Classification}]
This image is associated with the following caption: '!<INPUT 0>!'.\\
Does this combination exhibit any elements of hate speech? If so, which hate speech type does it belong to?\\
Choices: [NotHate, Racist, Sexist, Homophobe, Religion, OtherHate]
\end{tcolorbox}
\label{tab:prompt_hate2_cls_multi}
\end{center}

% \begin{table}[H]
% 	\centering
%     \begin{tabularx}{0.9\columnwidth}{ >{\raggedright\arraybackslash}X }
%     \toprule
%     \textbf{Task Description} \\
%     \midrule
%     \text{[Original text]: !<INPUT 0>!} \\
%     Identify the type of hate speech in the text following [original text], labeling it as either white-grievance, threatening, inferiority, stereotypical, incitement irony or other.\\
%     Restriction: Use only one word for your response.\\
%     \bottomrule
%     \end{tabularx}
%     \label{tab:prompt_hate_cls}
% \end{table}
% \par 
\noindent\textbf{Ideology \& Stance}\hspace{5px}The classification task of Ideology \& Stance primarily involves analyzing the ideological orientation of a given text or text-image pair, determining whether it aligns with liberal or conservative perspectives. \par 
\begin{center}
\begin{tcolorbox}[colback=orange!5!white,colframe=orange!55!black,width=0.95\columnwidth,title={Ideology Classification}]
\text{[Original text]: !<INPUT 0>!} \\
Analyze the political orientation reflected in the provided text [Original Text] and categorize it as either "Liberal" or "Conservative".\\
\text{[Original Text]: !<INPUT 0>!}\\
Note: Provide a response using only one of the two specified categories: "Liberal" or "Conservative".
\end{tcolorbox}
\label{tab:prompt_ideology_cls}
\end{center}

\begin{center}
\begin{tcolorbox}[colback=red!5!white,colframe=red!55!black,width=0.95\columnwidth,title={Multinodal Ideology Classification}]
This image is posted by a !<INPUT 0>! and is associated with the following caption: '!<INPUT 1>!'.\\
Question: What ideology does this !<INPUT 0>! belong to?\\
Choice: [left, center, right].
\end{tcolorbox}
\label{tab:prompt_ideology_cls_multi}
\end{center}
% \begin{table}[H]
% 	\centering
%     \begin{tabularx}{0.9\columnwidth}{ >{\raggedright\arraybackslash}X }
%     \toprule
%     \textbf{Task Description} \\
%     \midrule
%     Analyze the political orientation reflected in the provided text [Original Text] and categorize it as either "Liberal" or "Conservative".\\
%     \text{[Original Text]: !<INPUT 0>!}\\
%     Note: Provide a response using only one of the two specified categories: "Liberal" or "Conservative".\\
%     \bottomrule
%     \end{tabularx}
%     \label{tab:prompt_ideology_cls}
% \end{table}
% \par 
\textbf{Trustworthiness \& Social Bias}\hspace{5px}The classification task of Trustworthiness \& Social Bias primarily involves detecting the veracity of statements or determining whether they are exaggerated.
\begin{center}
\begin{tcolorbox}[colback=orange!5!white,colframe=orange!55!black,width=0.95\columnwidth,title={Trustworthiness Classification}]
Examine the given [Original Text] from an actual conversation to assess its truthfulness. Decide whether the statement is a 'truth' or a 'lie'.\\
\text{[Original Text]: !<INPUT 0>!}\\
Note: Please provide a brief response, choosing 'truth' or 'lie'.
\end{tcolorbox}
\label{tab:prompt_trust1_cls}
\end{center}

\begin{center}
\begin{tcolorbox}[colback=orange!5!white,colframe=orange!55!black,width=0.95\columnwidth,title={Trustworthiness Classification}]
Evaluate [Original Text] to find hyperbole. If there are exaggerated statements, over-the-top expressions, or intentional exaggeration, mark it as Hyperbole. Otherwise, label it as Not-Hyperbole.\\ 
\text{[Original Text]: !<INPUT 0>!}
\end{tcolorbox}
\label{tab:prompt_trust2_cls}
\end{center}
% \par 
\textbf{Social Factors}\hspace{5px}The classification task of social factors encompasses a variety of task types, such as determining whether a given statement is polite, whether the statement demonstrates empathy or complaint, assessing the level of intimacy in a conversation, and the selection and generation of hashtags.
\begin{center}
\begin{tcolorbox}[colback=orange!5!white,colframe=orange!55!black,width=0.95\columnwidth,title={Social Factors Classification}]
Examine the [Original Text] for its overall tone, determining its classification as 'polite' or 'impolite'.\\
\text{[Original Text]: !<INPUT 0>!}\\
Instruction: Provide a straightforward response, selecting 'polite' or 'impolite'.
\end{tcolorbox}
\label{tab:prompt_social1_cls}
\end{center}

\begin{center}
\begin{tcolorbox}[colback=orange!5!white,colframe=orange!55!black,width=0.95\columnwidth,title={Social Factors Classification}]
Review the supplied [Original Text] to decide if it shows signs of 'empathy' or the absence thereof.\\
\text{[Original Text]: !<INPUT 0>!}\\
Obligation: Give a terse verdict, choosing between 'empathy' or 'not empathy'.
\end{tcolorbox}
\label{tab:prompt_social2_cls}
\end{center}

\begin{center}
\begin{tcolorbox}[colback=orange!5!white,colframe=orange!55!black,width=0.95\columnwidth,title={Social Factors Classification}]
Evaluate the given [Original Text] to ascertain whether it falls under the classification of 'complaint' or 'not complaint'.\\
\text{[Original Text]: !<INPUT 0>!}\\
Instruction: Provide a brief and clear decision, opting for either 'complaint' or 'not complaint' as the suitable categorization.
\end{tcolorbox}
\label{tab:prompt_social3_cls}
\end{center}

\begin{center}
\begin{tcolorbox}[colback=orange!5!white,colframe=orange!55!black,width=0.95\columnwidth,title={Social Factors Classification}]
Determine the intimacy level in the provided [Original Text]. Classify it as Very-intimate, Intimate, Somewhat-intimate, Not-very-intimate, Not-intimate, or Not-intimate-at-all using the following criteria.\\
criteria:\\
Very-intimate: the text involves a deeply personal or private matter, elicits a strong emotional response, or requires sharing sensitive information. \\
Intimate: the text involve sharing personal preferences, experiences, or opinions that go beyond surface-level topics.\\
Somewhat-intimate: the text touches on personal matters to some extent but is not as deep.\\
Not-very-intimate: the text discusses general or non-personal topics. \\
Not-intimate: the text is unrelated to personal matters or feelings. \\
Not-intimate-at-all: the text is entirely unrelated to personal matters and is more factual or transactional.\\
\text{[Original Text]: !<INPUT 0>!}\\
Constraint: Provide a single-word response.
\end{tcolorbox}
\label{tab:prompt_social4_cls}
\end{center}

\begin{center}
\begin{tcolorbox}[colback=red!5!white,colframe=red!55!black,width=0.95\columnwidth,title={Multimodal Social Factors Classification}]
This image and the following caption are from the same user: '!<INPUT 0>!'\\
Is the user likely to be male or female?\\
Pick your answer from [male, female].
\end{tcolorbox}
\label{tab:prompt_social1_cls_multi}
\end{center}

\begin{center}
\begin{tcolorbox}[colback=red!5!white,colframe=red!55!black,width=0.95\columnwidth,title={Multimodal Social Factors Classification}]
This image is associated with the following caption by an Instagram user.\\
caption: !<INPUT 0>!\\
Which of the following hashtags BEST describes this post?\\
Choices: [!<INPUT 1>!, !<INPUT 2>!, !<INPUT 3>!, !<INPUT 4>!]\\
Constraints: only choose ONE hashtag from the Choice, and \# should be included.
\end{tcolorbox}
\label{tab:prompt_social2_cls_multi}
\end{center}

\subsubsection{Application Module}
As discussed in~\S\ref{subsec:applying}, the primary function of the Application Module is to interpret the ground truth labels of a given text. \\ 
\textbf{Emotion}\hspace{5px}The task within the "Application Module" related to emotions involves extracting the trigger that elicits a specific emotion, given the ground truth label of a provided text.
\begin{center}
\begin{tcolorbox}[colback=orange!5!white,colframe=orange!55!black,width=0.95\columnwidth,title={Emotion Trigger Extraction}]
The provided statement conveys a !<INPUT 1>! emotion. Kindly identify the stimuli that evoke this emotion.\\
\text{[sentence]: !<INPUT 0>!}
\end{tcolorbox}
\label{tab:prompt_emotion_gen}
\end{center}
\textbf{Humor}\hspace{5px}The task within the "Application Module" related to humor is to provide corresponding explanations for statements labeled as humor in the ground truth data.
\begin{center}
\begin{tcolorbox}[colback=orange!5!white,colframe=orange!55!black,width=0.95\columnwidth,title={Humor Explanation}]
Consideration is given to the sentence being categorized as humor. Please elucidate the reasoning behind this classification.\\
\text{[sentence]: !<INPUT 0>!}
\end{tcolorbox}
\label{tab:prompt_humor_gen}
\end{center}
\textbf{Hate Speech \& Toxicity}\hspace{5px}The task within the "Application Module" related to Hate Speech is aimed at providing explanations for texts classified as a certain type of Hate Speech.
\begin{center}
\begin{tcolorbox}[colback=orange!5!white,colframe=orange!55!black,width=0.95\columnwidth,title={Hate Speech Explanation}]
The sentences below are flagged for !<INPUT 1>! concerns. Please provide a concise explanation. \\
\text{[sentence]: !<INPUT 0>!}
\end{tcolorbox}
\label{tab:prompt_hate_gen}
\end{center}
\textbf{Ideology \& Stance}\hspace{5px}The task within the "Application Module" regarding Ideology is to furnish corresponding explanations for texts categorized under a certain ideology (liberal or conservative).
\begin{center}
\begin{tcolorbox}[colback=orange!5!white,colframe=orange!55!black,width=0.95\columnwidth,title={Ideology Explanation}]
The following sentence suggests a perspective aligned with !<INPUT 1>!; Please provide a concise explanation.\\
\text{[sentence]: !<INPUT 0>!}
\end{tcolorbox}
\label{tab:prompt_ideology_gen}
\end{center}
\textbf{Trustworthiness \& Social Bias}\hspace{5px}The task of assessing trustworthiness and bias within the "Application Module" involves analyzing two given texts to determine which one exhibits greater bias.
\begin{center}
\begin{tcolorbox}[colback=orange!5!white,colframe=orange!55!black,width=0.95\columnwidth,title={Social Bias Explanation}]
Here we have two sentences. Kindly explain in a brief manner why !<INPUT 2>! is short.\\
\text{[sentence]: !<INPUT 0>!}\\
\text{[sentence]: !<INPUT 1>!}
\end{tcolorbox}
\label{tab:prompt_trustworth_gen}
\end{center}

\noindent\textbf{Social Factors}\hspace{5px}The social factor task within the application module consists of tasks to explain a user's domain or personality given a text-image pair post by the user.
\begin{center}
\begin{tcolorbox}[colback=red!5!white,colframe=red!55!black,width=0.95\columnwidth,title={Multimodal Social Factors Explanation}]
This image is linked with the following caption provided by a user.\\
Caption: !<INPUT 0>!\\
What is the user's professional field? Please explain in one sentence.
\end{tcolorbox}
\label{tab:prompt_socialfactor1_gen_multi}
\end{center}

\begin{center}
\begin{tcolorbox}[colback=red!5!white,colframe=red!55!black,width=0.95\columnwidth,title={Multimodal Social Factors Explanation}]
This image is associated with the following caption by an Instagram user.\\
caption: !<INPUT 0>!\\
What's the personality of this user according to the post?\\
Constraints: First give the personality and explain it in one sentence.
\end{tcolorbox}
\label{tab:prompt_socialfactor2_gen_multi}
\end{center}

\subsubsection{Analysis Module}
\textbf{Figurative Language}\hspace{5px}The task of Figurative Language in the Analysis Module involves enabling the model to analyze whether a text contains figurative language without the aid of known labels and to provide corresponding interpretations.
\begin{center}
\begin{tcolorbox}[colback=orange!5!white,colframe=orange!55!black,width=0.95\columnwidth,title={Figurative Language Analysis}]
Interpret the metaphorical or symbolic use of language in the following hypothesis in a single sentence. \\
\text{[Hypothesis]: !<INPUT 0>!}
\end{tcolorbox}
\label{tab:prompt_figurative_ana}
\end{center}

\noindent\textbf{Emotion}\hspace{5px}The task of Emotion in the Analysis Module asks the model to generate the emotion or sentiment directly without any labels given.
\begin{center}
\begin{tcolorbox}
[colback=red!5!white,colframe=red!55!black,width=0.95\columnwidth,title={Multimodal Emotion Analysis}]
This image is associated with the following caption: '!<INPUT 0>!'.\\
What fine-grained emotion does this combination convey?
\end{tcolorbox}
\label{tab:prompt_emo_ana_multi}
\end{center}

\noindent\textbf{Hate Speech \& Toxicity}\hspace{5px} The task of Hate Speech \& Toxicity in the Analysis Module asks the model to identify whether the text-image pair contains any hate speech directly without any labels given.
\begin{center}
\begin{tcolorbox}[colback=red!5!white,colframe=red!55!black,width=0.95\columnwidth,title={Multimodal Hate Speech Analysis}]
This image is associated with the following caption: '!<INPUT 0>!'.\\
Does this combination exhibit any elements of hate speech? If so, which hate speech type does it belong to?
\end{tcolorbox}
\label{tab:prompt_hatespeech_ana_multi}
\end{center}

\noindent\textbf{Social Factors}\hspace{5px}The task of Social Factors in the Analysis Module asks the model to identify the gender of the user given the text-image pair without labels given.
\begin{center}
\begin{tcolorbox}[colback=red!5!white,colframe=red!55!black,width=0.95\columnwidth,title={Multimodal Social Factors Analysis}]
Determine the gender of the user given the following information.\\
This image and the following caption are from the same user: '!<INPUT 0>!'
\end{tcolorbox}
\label{tab:prompt_socialfactors_ana_multi}
\end{center}

\subsubsection{Evaluation Module}
\textbf{Ideology \& Stance}\hspace{5px}The task of Stance \& Ideology in the Evaluation Module asks the model to identify the stance of the user given the text-image pair without labels given.
\begin{center}
\begin{tcolorbox}[colback=red!5!white,colframe=red!55!black,width=0.95\columnwidth,title={Multimodal Ideolog \& Stance Evaluation}]
This image is associated with the following caption: '!<INPUT 0>!'.\\
It is posted by a politician. What ideology does the politician belong to?
\end{tcolorbox}
\label{tab:prompt_stance_eva_multi}
\end{center}

\noindent\textbf{Misinformation}\hspace{5px}The task of Misinformation within the Evaluation Module is aimed at interpreting the deep-seated implications of news headlines.
\begin{center}
\begin{tcolorbox}[colback=orange!5!white,colframe=orange!55!black,width=0.95\columnwidth,title={Misinformation Evaluation}]
Deduce the underlying implication of the news headline below. Provide a brief response, similar in style to 'some masks are better than others.' \\
\text{[HEADLINE]: !<INPUT 0>!}
\end{tcolorbox}
\label{tab:prompt_misinfo_eva}
\end{center}

\begin{center}
\begin{tcolorbox}[colback=red!5!white,colframe=red!55!black,width=0.95\columnwidth,title={Multimodal Misinformation Evaluation}]
This news is accompanied by an image and has the following description:\\
!<INPUT 0>!\\
\\
Can the authenticity of this article be verified?
\end{tcolorbox}
\label{tab:prompt_misinfo_eva_multi}
\end{center}

\noindent\textbf{Trustworthiness \& Social Bias}\hspace{5px}The task of Trustworthiness within the Evaluation Module aims to detect rumors and provide corresponding explanations.
\begin{center}
\begin{tcolorbox}[colback=orange!5!white,colframe=orange!55!black,width=0.95\columnwidth,title={Trustworthiness Evaluation}]
The sentence implies that it embodies the quality of being !<INPUT 1>!. Kindly offer an explanation for the basis of this statement.\\
\text{[sentence]: !<INPUT 0>!}
\end{tcolorbox}
\label{tab:prompt_trust_eva}
\end{center}
\textbf{Detoxifying Content}\hspace{5px}The task of "Detoxifying Content" within the Evaluation Module aims to rewrite hate speech, reducing its toxicity.
\begin{center}
\begin{tcolorbox}[colback=orange!5!white,colframe=orange!55!black,width=0.95\columnwidth,title={Detoxifying Content}]
Revise this sentence to make it more positive while keeping its original intent as much as possible.\\
\text{[sentence]: !<INPUT 0>!}
\end{tcolorbox}
\label{tab:prompt_detoxify_eva}
\end{center}
\textbf{Depolarizing Language}\hspace{5px}The task of Depolarizing Language in the Evaluation Module is aimed at depolarizing ideological discourse.
\begin{center}
\begin{tcolorbox}[colback=orange!5!white,colframe=orange!55!black,width=0.95\columnwidth,title={Depolarizing Language}]
Rewording the following sentence to make it less polarizing.\\
\text{[sentence]: !<INPUT 0>!}
\end{tcolorbox}
\label{tab:prompt_depolarize_eva}
\end{center}
\subsubsection{Creation Module}
% \textbf{Invert Opinion}\hspace{5px}
\textbf{Reverse Ideology}\hspace{5px}The task of Reverse Ideology in the Creation Module involves providing the model with a text characterized by a specific ideology (either liberal or conservative) and prompting the model to produce statements on the same topic that reflect the opposite ideology.
\begin{center}
\begin{tcolorbox}[colback=orange!5!white,colframe=orange!55!black,width=0.95\columnwidth,title={Depolarizing Language}]
Reframe this !<INPUT 1>! speech from a !<INPUT 2>! perspective, ensuring the core theme remains the same.\\ 
\text{[sentence]: !<INPUT 0>!}
\end{tcolorbox}
\label{tab:prompt_reverse_create}
\end{center}
\textbf{Social Factors}\hspace{5px}The task of Social Factors in the Creation Module involves providing the model with a text-image pair and prompting the model to generate three hashtags that best summarize the post.
\begin{center}
\begin{tcolorbox}[colback=red!5!white,colframe=red!55!black,width=0.95\columnwidth,title={Multimodel Hashtag Generation}]
This image is associated with the following caption by an Instagram user.\\
Caption: !<INPUT 0>!\\
Try to generate no more than 3 hashtags that best fit this post. \\
Constraints: the hashtags should begin with \#.\\
Output Format: \#hashtag\_1, \# hashtag\_2, \# hashtag\_3
\end{tcolorbox}
\label{tab:hashtag_gen_create_multi}
\end{center}

\section{Training Details}\label{appendix:trainsetting}
\subsection{Computational resources}
All of our experiments were conducted on an Ubuntu 22.04.3 machine installed with NVIDIA RTX 3090 and 4090 GPUs. The Python packages used in our experiments include Pytorch 2.1.1, Transformers 4.33.0, and deepspeed 0.11.1.
\subsection{Details on large language model instruction tuning}
As mentioned in~\S\ref{subsec:implement}, we employ the QLoRA method~\citep{dettmers2023qlora} with FastChat~\citep{zheng2023judging} for language model tuning. The specific settings for the hyper-parameters are presented in Table~\ref{tab:appendix-para-language}.
\begin{table}[h!]
	\centering
	%\resizebox{1\columnwidth}{!}{
	\begin{tabular}{lc}
		\toprule[1.1pt]
		\textbf{Hyper-parameters} & \textbf{Value} \\
		\midrule
		lora\_r & 128 \\ 
		lora\_alpha & 256 \\
		per\_device\_train\_batch\_size & 8 \\
		gradient\_accumulation\_steps & 2 \\
		learning\_rate & 2e-5 \\
		weight\_decay& 0. \\   
		warmup\_ratio & 0.05 \\
		lr\_scheduler\_type & cosine \\ 
		tf32 & True \\
		model\_max\_length & 2048 \\
		q\_lora & True \\
		flash\_attn & True \\ 
		\bottomrule[1.1pt]
	\end{tabular}
	%}
	\caption{Hyper-parameters of Language Model Tuning}
	\label{tab:appendix-para-language}
\end{table}

\subsection{Details on Q-former instruction tuning}
As mentioned in~\S\ref{subsec:implement}, we tuned our connection module following the pipeline of LAVIS~\cite{li-etal-2023-lavis}. The specific settings for the hyperparameters are presented in Table~\ref{tab:appendix-para-qformer}.
\begin{table}[h!]
\centering
\begin{tabular}{lc}
\toprule[1.1pt]
\textbf{Hyper-parameters}   & \textbf{Value}                      \\ \midrule
init\_lr           & 3e-5                       \\
min\_lr            & 1e-5                       \\
lr\_sched          & linear\_warmup\_cosine\_lr \\
weight\_decay      & 0.02                       \\
max\_epoch         & 3                          \\
batch\_size\_train & 1                          \\
batch\_size\_eval  & 1                          \\
num\_workers       & 1                          \\
freeze\_vit        & True                       \\ \bottomrule[1.1pt]
\end{tabular}
\caption{Hyperparameters of Connection Module Tuning.}
\label{tab:appendix-para-qformer}
\end{table}

\section{Experiment Results on Each Dataset}\label{sec:appendix-results}
\subsection{Textual Datasets}
Experiment results on each dataset in textual tasks are shown in Table~\ref{tab:appendix-text-cls} and Table~\ref{tab:appendix-text-gen}.

\subsection{Multimodal Datasets}
Experiment results on each dataset in multimodal tasks are shown in Table~\ref{tab:appendix-mult-cls} and Table~\ref{tab:appendix-mult-gen}.

\begin{table*}[t!]
\centering
% \resizebox{\textwidth}{!}{%
\begin{tabular}{lcccc}
\hline
                       & SoMeLVLM        & Vicuna       & Llama2          & Chatglm2        \\ 
Datasets                & Accuracy        & Accuracy     & Accuracy        & Accuracy        \\ \hline
Twitter\_emotion       & \textbf{80.66} & 35.86       & 40.54          & {\ul 41.20}     \\
hahackathon\#is\_humor & {\ul 60.47}    & 41.08       & \textbf{61.31} & 36.94          \\
tweet\_irony           & \textbf{61.70}  & 47.08       & {\ul 53.77}    & 52.05          \\
misinfo\_cancer        & \textbf{70.38} & {\ul 59.23} & 41.11          & 47.21          \\
latent\_hatred         & \textbf{22.20}  & 11.94       & 12.84          & {\ul 14.67}    \\
media\_ideology        & \textbf{45.23} & 34.15       & {\ul 37.77}    & 30.08          \\
hypo-l                 & 43.52          & 36.60        & {\ul 59.21}    & \textbf{68.44} \\
hayati\_politeness     & \textbf{89.68} & 70.63       & 49.69          & {\ul 83.43}    \\
question intimacy       & \textbf{21.09} & {\ul 14.73} & 13.53          & 13.03          \\ \hline
\end{tabular}%
% }
\caption{Classification results on each dataset in the textual experiment.}
\label{tab:appendix-text-cls}
\end{table*}

\begin{table*}[t!]
\centering
\resizebox{\textwidth}{!}{%
\begin{tabular}{lcccccccccccc}
\hline
                                             & \multicolumn{3}{c}{SoMeLVLM}                    & \multicolumn{3}{c}{Vicuna}                & \multicolumn{3}{c}{Llama2}       & \multicolumn{3}{c}{Chatglm2} \\
Dataset                                      & BLEU           & ROUGE          & Score         & BLEU        & ROUGE       & Score         & BLEU        & ROUGE & Score      & BLEU     & ROUGE   & Score   \\ \hline
twitter\_emotion\_EXP                        & \textbf{26.96} & \textbf{51.88} & \textbf{3.63} & {\ul 7.97}  & {\ul 31.31} & {\ul 3.23}    & 4.25        & 23.50 & 2.99       & 6.60     & 29.47   & 3.05    \\
hahackathon\#is\_humor\_EXP                  & \textbf{13.81} & \textbf{42.84} & \textbf{3.38} & {\ul 10.49} & {\ul 36.21} & {\ul 3.24}    & 6.36        & 28.37 & 2.48       & 8.98     & 34.49   & 2.37    \\
tweet\_irony\_EXP                            & \textbf{23.77} & \textbf{45.42} & \textbf{3.02} & 8.03        & {\ul 31.55} & 2.57          & {\ul 10.39} & 31.32 & {\ul 2.73} & 7.20     & 29.07   & 2.06    \\
contextual-abuse\#IdentityDirectedAbuse\_EXP & \textbf{18.10} & \textbf{43.36} & \textbf{3.55} & {\ul 6.49}  & {\ul 30.80} & {\ul 3.46}    & 1.69        & 17.72 & 1.96       & 4.24     & 27.19   & 2.60    \\
contextual-abuse\#PersonDirectedAbuse\_EXP   & \textbf{18.56} & \textbf{45.38} & \textbf{3.72} & {\ul 6.86}  & {\ul 30.22} & {\ul 3.62}    & 1.38        & 15.28 & 1.55       & 4.50     & 27.53   & 2.71    \\
implicit-hate\#explicit\_hate\_EXP           & \textbf{20.76} & \textbf{47.49} & \textbf{3.85} & {\ul 8.09}  & {\ul 33.11} & {\ul 3.83}    & 2.11        & 19.02 & 2.09       & 4.77     & 28.90   & 3.42    \\
implicit-hate\#implicit\_hate\_EXP           & \textbf{14.87} & \textbf{39.78} & {\ul 3.52}    & {\ul 6.82}  & {\ul 31.37} & \textbf{3.61} & 1.78        & 17.43 & 1.97       & 4.23     & 28.33   & 2.94    \\
latent\_hatred\_EXP                          & \textbf{13.89} & \textbf{39.51} & {\ul 3.58}    & {\ul 6.08}  & {\ul 30.72} & \textbf{3.62} & 1.99        & 17.60 & 2.13       & 4.75     & 28.29   & 3.02    \\
media\_ideology\_EXP                         & \textbf{14.60} & \textbf{39.49} & \textbf{3.43} & {\ul 9.36}  & {\ul 32.78} & {\ul 3.41}    & 4.75        & 25.01 & 2.78       & 6.59     & 29.94   & 2.86    \\
rumor\#rumor\_bool\_EXP                      & \textbf{12.37} & \textbf{39.06} & \textbf{3.59} & {\ul 9.70}  & {\ul 34.13} & {\ul 3.13}    & 4.73        & 26.54 & 2.82       & 9.25     & 34.35   & 2.73    \\
contextual-abuse\#IdentityDirectedAbuse\_EXP & \textbf{28.11} & \textbf{48.68} & \textbf{3.00} & {\ul 11.00} & {\ul 28.47} & {\ul 2.60}    & 1.57        & 11.54 & 1.23       & 6.50     & 22.85   & 2.00    \\
contextual-abuse\#PersonDirectedAbuse\_EXP   & \textbf{29.64} & \textbf{49.39} & \textbf{3.08} & {\ul 11.37} & {\ul 28.21} & {\ul 2.66}    & 1.67        & 12.13 & 1.34       & 6.62     & 23.25   & 2.08    \\
implicit-hate\#explicit\_hate\_EXP           & \textbf{22.98} & \textbf{43.78} & \textbf{2.50} & {\ul 7.15}  & {\ul 23.76} & {\ul 2.07}    & 0.80        & 9.24  & 0.90       & 5.92     & 22.63   & 1.74    \\
implicit-hate\#implicit\_hate\_EXP           & \textbf{27.77} & \textbf{49.18} & \textbf{2.97} & {\ul 12.21} & {\ul 31.38} & {\ul 2.69}    & 1.21        & 10.85 & 1.07       & 8.30     & 26.94   & 2.18    \\
media\_ideology\_EXP                         & \textbf{23.54} & \textbf{45.47} & \textbf{3.28} & {\ul 22.31} & {\ul 42.72} & {\ul 3.26}    & 8.40        & 26.72 & 2.21       & 13.33    & 35.66   & 2.80    \\
media\_ideology\_EXP                         & \textbf{44.09} & \textbf{61.96} & \textbf{3.41} & {\ul 33.40} & {\ul 51.76} & {\ul 2.981}   & 20.54       & 38.06 & 2.04       & 21.91    & 42.27   & 2.80   \\ \hline
\end{tabular}%
}
\caption{Generation results on each dataset in the textual experiment.}
\label{tab:appendix-text-gen}
\end{table*}

\begin{table*}[t!]
\centering
\renewcommand\arraystretch{1.2}
\resizebox{\textwidth}{!}{
\begin{tabular}{lcccccccccccc}
\hline
                & \multicolumn{2}{c}{SoMeLVLM} & \multicolumn{2}{c}{Instructblip$_V$}                & \multicolumn{2}{c}{Instructblip$_F$}                & \multicolumn{2}{c}{Blip2}                          & \multicolumn{2}{c}{Llava}                          & \multicolumn{2}{c}{Minigpt4}                       \\
Datasets        & Acc*        & Acc        & \multicolumn{1}{c}{Acc*} & \multicolumn{1}{c}{Acc} & \multicolumn{1}{c}{Acc*} & \multicolumn{1}{c}{Acc} & \multicolumn{1}{c}{Acc*} & \multicolumn{1}{c}{Acc} & \multicolumn{1}{c}{Acc*} & \multicolumn{1}{c}{Acc} & \multicolumn{1}{c}{Acc*} & \multicolumn{1}{c}{Acc} \\ \hline
4chans          & 75.00       & 75.00      & 55.49                    & 50.50                   & 57.47                    & 56.75                   & 56.00                    & 56.00                   & 79.49                    & 15.50                   & 66.14                    & 41.50                   \\
MMHS            & 67.40       & 67.40      & 22.01                    & 13.60                   & 31.65                    & 31.40                   & 34.00                    & 34.00                   & 29.53                    & 11.40                   & 18.08                    & 9.40                    \\
FakeNewsNet     & 82.60       & 82.60      & 47.55                    & 13.60                   & 80.78                    & 79.00                   & 80.60                    & 80.60                   & 84.67                    & 25.40                   & 65.30                    & 54.20                   \\
hatefulmemes    & 75.80       & 75.80      & 50.13                    & 39.60                   & 63.50                    & 58.80                   & 67.20                    & 67.20                   & 56.25                    & 3.60                    & 55.33                    & 21.80                   \\
MVSA\_single    & 76.05       & 76.05      & 58.27                    & 53.88                   & 70.09                    & 69.62                   & 70.07                    & 70.07                   & 62.50                    & 4.43                    & 57.39                    & 29.27                   \\
MVSA\_multiple  & 67.60       & 67.60      & 59.28                    & 55.60                   & 65.12                    & 64.60                   & 64.40                    & 64.40                   & 65.21                    & 3.00                    & 62.31                    & 33.40                   \\
PAN             & 69.00       & 69.00      & 68.92                    & 55.00                   & 64.92                    & 64.40                   & 64.80                    & 64.80                   & 54.37                    & 11.20                   & 56.71                    & 41.40                   \\
TumEmo          & 48.19       & 48.10      & 46.50                    & 37.80                   & 42.70                    & 40.45                   & 40.04                    & 40.04                   & 33.43                    & 22.36                   & 40.19                    & 25.81                   \\
tweet\_leg      & 83.45       & 64.36      & 65.25                    & 48.94                   & 62.05                    & 54.79                   & 55.32                    & 55.32                   & 66.67                    & 2.12                    & 50.00                    & 9.04                    \\
tweet\_cele     & 58.24       & 41.41      & 37.84                    & 32.81                   & 41.41                    & 32.03                   & 50.78                    & 50.78                   & 25.00                    & 0.78                    & 30.56                    & 8.59                    \\
hashtag\_choice & 99.38       & 65.64      & 91.30                    & 26.64                   & 98.00                    & 82.88                   & 99.13                    & 97.25                   & 90.91                    & 2.11                    & 71.57                    & 30.87                   \\ \hline
\end{tabular}
}
\caption{Classification results on each dataset in the multimodal experiment.}
\label{tab:appendix-mult-cls}
\end{table*}

\begin{table*}[t!]
\centering
\renewcommand\arraystretch{1.2}
\resizebox{\textwidth}{!}{
\begin{tabular}{lcccccccccccccccccc}
\hline
\multicolumn{1}{c}{} & \multicolumn{3}{c}{SoMeLVLM} & \multicolumn{3}{c}{Instructblip\(_V\)} & \multicolumn{3}{c}{Instructblip\(_F\)} & \multicolumn{3}{c}{Blip2} & \multicolumn{3}{c}{Llava} & \multicolumn{3}{c}{Minigpt4} \\
Datasets             & BLEU     & ROUGE    & GPT    & BLEU       & ROUGE      & GPT       & BLEU       & ROUGE      & GPT       & BLEU   & ROUGE   & GPT    & BLEU   & ROUGE   & GPT    & BLEU    & ROUGE    & GPT     \\ \hline
4chans\_EXP          & 27.42    & 49.76    & 3.33   & 0.74       & 3.34       & 1.60      & 0.42       & 4.23       & 1.51      & 1.29   & 5.18    & 1.63   & 0.46   & 6.06    & 1.27   & 0.54    & 9.91     & 3.15    \\
hatefulmemes\_EXP    & 33.37    & 48.60    & 2.83   & 0.53       & 3.17       & 2.37      & 0.23       & 3.39       & 2.63      & 0.15   & 1.10    & 2.13   & 0.39   & 5.07    & 1.29   & 0.36    & 9.19     & 1.95    \\
MMHS\_EXP            & 32.34    & 40.68    & 3.49   & 0.69       & 2.87       & 1.47      & 0.07       & 0.75       & 2.07      & 0.41   & 0.46    & 1.76   & 0.22   & 2.43    & 1.14   & 0.38    & 7.41     & 1.90    \\
FakeNewsNet\_EXP     & 24.06    & 43.22    & 2.94   & 1.09       & 6.21       & 2.84      & 0.05       & 0.81       & 2.85      & 0.02   & 1.89    & 2.72   & 0.00   & 0.01    & 0.81   & 0.69    & 12.15    & 2.18    \\
PAN\_EXP             & 35.42    & 61.05    & 3.48   & 0.39       & 6.21       & 1.00      & 1.17       & 22.16      & 2.88      & 0.15   & 21.39   & 3.17   & 1.47   & 9.81    & 1.54   & 0.42    & 23.95    & 1.64    \\
hashtag\_gen         & 2.94     & 8.51     & 1.10   & 0.95       & 1.07       & 0.80      & 0.60       & 1.78       & 1.14      & 1.52   & 0.53    & 1.12   & 1.96   & 2.43    & 1.08   & 0.85    & 4.97     & 1.06    \\
domain\_explain      & 10.25    & 31.94    & 3.35   & 0.57       & 13.27      & 1.67      & 1.29       & 15.80      & 2.09      & 0.92   & 13.98   & 1.71   & 1.77   & 19.35   & 2.03   & 1.78    & 20.57    & 1.83    \\
personality\_explain & 9.33     & 29.98    & 3.50   & 1.62       & 15.52      & 2.40      & 1.56       & 18.65      & 2.34      & 0.45   & 12.06   & 1.53   & 2.35   & 19.62   & 2.54   & 1.73    & 19.30    & 1.85    \\
MVSA\_multiple\_EXP  & 42.91    & 60.58    & 3.80   & 1.15       & 9.64       & 2.24      & 0.23       & 19.26      & 3.65      & 0.22   & 22.74   & 3.82   & 0.88   & 6.73    & 1.61   & 0.71    & 11.63    & 2.79    \\
MVSA\_single\_EXP    & 39.38    & 59.12    & 3.78   & 0.85       & 6.60       & 1.88      & 0.23       & 17.31      & 3.36      & 0.21   & 21.43   & 3.59   & 0.83   & 6.53    & 1.51   & 0.68    & 11.87    & 2.55    \\
TumEmo\_EXP          & 30.66    & 41.92    & 3.03   & 0.56       & 5.54       & 1.75      & 0.39       & 4.49       & 2.09      & 0.06   & 0.28    & 1.88   & 0.21   & 3.95    & 0.64   & 0.26    & 8.93     & 1.79    \\
tweet\_cele\_EXP     & 19.02    & 37.45    & 2.75   & 0.41       & 3.53       & 1.14      & 0.86       & 8.06       & 1.07      & 0.24   & 2.78    & 2.23   & 0.76   & 6.40    & 0.54   & 0.29    & 13.26    & 0.59    \\
tweet\_leg\_EXP      & 29.14    & 44.62    & 3.82   & 0.79       & 6.24       & 1.93      & 0.69       & 8.65       & 1.99      & 0.26   & 5.92    & 2.42   & 1.44   & 11.06   & 1.66   & 0.34    & 12.10    & 1.75    \\
domain\_ood          & 10.41    & 31.85    & 3.38   & 0.49       & 11.73      & 1.62      & 1.26       & 15.11      & 2.04      & 0.88   & 13.85   & 1.66   & 2.07   & 20.23   & 1.97   & 1.89    & 20.88    & 1.74    \\
personality\_ood     & 9.95     & 30.20    & 3.52   & 1.79       & 16.33      & 2.53      & 1.75       & 18.70      & 2.29      & 0.41   & 11.89   & 1.56   & 2.51   & 19.97   & 2.58   & 2.07    & 20.57    & 1.95    \\ \hline
\end{tabular}
}
\caption{Generation results on each dataset in the multimodal experiment.}
\label{tab:appendix-mult-gen}
\end{table*}

\end{document}